%% file: root.tex
\documentclass[letterpaper, 10 pt, conference]{ieeeconf}  % Comment this line out if you need a4paper
\usepackage[utf8]{inputenc}

\IEEEoverridecommandlockouts                              % This command is only needed if 
\usepackage{amsmath}
\usepackage{amssymb}
\usepackage{mathrsfs}
\usepackage{graphicx}

\usepackage{balance} % for balancing columns on the final page
\usepackage{cuted}          % for \begin{strip}
\usepackage{capt-of}        % to use \captionof
\usepackage{subcaption}
\usepackage{algorithm}
\usepackage{algpseudocode}   % for \State, \Function, \For, etc.
\usepackage[rightComments,indLines,italicComments=false,commentColor=gray]{algpseudocodex}
\algrenewcommand\alglinenumber[1]{\scriptsize #1} % optional: line numbers smaller
\algrenewcommand\algorithmicindent{0.8em}
\usepackage{xcolor}
\usepackage{cite}
\usepackage{hyperref}

\newcommand{\robot}[1]{R^{#1}}
\newcommand{\robotset}{\mathcal{R}}
\newcommand{\robcfg}[1]{q_{\robot{}}^{#1}}
\newcommand{\robstartcfgset}{Q_{\robot{}, \text{start}}}
\newcommand{\robgoalcfgset}{Q_{\robot{}, \text{goal}}}
\newcommand{\robcfgspace}[1]{\mathcal{Q}_{\robot{}}^{#1}}
\newcommand{\robtraj}[2]{\tau^{#1}_{#2}}
\newcommand{\robtrajset}[1]{\mathcal{T}^{#1}}

\newcommand{\obj}[1]{\mathit{o}_{#1}}
\newcommand{\objset}{\mathcal{O}}
\newcommand{\objtraj}[1]{\tau^{#1}_{\text{obj}}}

\newcommand{\objcfg}[1]{q_{\obj{}}^{#1}}
\newcommand{\objgoalcfg}[1]{q_{\obj{}, \text{goal}}^{#1}}
\newcommand{\objcfgspace}[1]{\mathcal{Q}_{\obj{}}^{#1}}

\newcommand{\contact}[2]{k^{#1}_{#2}}
\newcommand{\contactset}[1]{\mathcal{K}^{#1}}
\newcommand{\contactspace}{\mathscr{K}}
\newcommand{\tok}[2]{{c}^{#1}_{#2}}

\newcommand{\image}[1]{\mathcal{I}^{#1}}

\newcommand{\mprimgen}[1]{\mathcal{G}^{#1}}

\newcommand{\gco}{\textsc{GCo}}

\newcommand{\gcodc}{$\textsc{GCo}_{DC}$}
\newcommand{\gcocc}{$\textsc{GCo}_{CC}$}
\newcommand{\gcoct}{$\textsc{GCo}_{C\robtrajset{}{}}$}

\newcommand{\gspi}{\textsc{Gspi}}

\title{\LARGE \bf
Collaborative Multi-Robot Non-Prehensile Manipulation \\ via Flow Matching Co-Generation}

\author{Yorai Shaoul$^{1}$, Zhe Chen$^{2,*}$, Naveed Gul Mohamed$^{2,*}$, Federico Pecora$^{2}$, Maxim Likhachev$^{1}$, Jiaoyang Li$^{1}$% <-this % stops a space
\thanks{{\scriptsize $^{1}$Carnegie Mellon University,
        {\tt \{yshaoul,maxim,jiaoyanl\}@cs.cmu.edu}}}%
\thanks{{\scriptsize $^{2}$Amazon Robotics,
        \tt \{zhecm,naveedg,fpecora\}@amazon.com}}%
\thanks{{\scriptsize During part of the work, Yorai interned at Amazon Robotics. *Equal contribution.}}% <-this % stops a space
}

\begin{document}

\maketitle
\thispagestyle{empty}
\pagestyle{empty}
\input{figures/fig_teaser}

%%%%%%%%%%%%%%%%%%%%%%%%%%%%%%%%%%%%%%%%%%%%%%%%%%%%%%%%%%%%%%%%%%%%%%%%%%%%%%%%
\begin{abstract}

Coordinating a team of robots to reposition multiple objects in cluttered environments requires reasoning jointly about where robots should establish contact, how to manipulate objects once contact is made, and how to navigate safely and efficiently at scale. Prior approaches typically fall into two extremes--either learning the entire task or relying on privileged information and hand-designed planners--both of which struggle to handle diverse objects in long-horizon tasks. To address these challenges, we present a unified framework for collaborative multi-robot, multi-object non-prehensile manipulation that integrates flow matching co-generation with anonymous multi-robot motion planning. Within this framework, a generative model co-generates contact formations and manipulation trajectories from visual observations, while a novel motion planner conveys robots at scale. Crucially, the same planner also supports coordination at the object level, assigning manipulated objects to larger target structures and thereby unifying robot- and object-level reasoning within a single algorithmic framework. Experiments in challenging simulated environments demonstrate that our approach outperforms baselines in both motion planning and manipulation tasks, highlighting the benefits of generative co-design and integrated planning for scaling collaborative manipulation to complex multi-robot, multi-object settings. 
Visit \href{https://gco-paper.github.io}{\texttt{gco-paper.github.io}} for demonstrations.

\end{abstract}

%%%%%%%%%%%%%%%%%%%%%%%%%%%%%%%%%%%%%%%%%%%%%%%%%%%%%%%%%%%%%%%%%%%%%%%%%%%%%%%%
\input{tex/introduction}
\input{tex/problem}
\input{tex/related}
\input{tex/method}
\input{tex/experiments}
% \input{figures/fig_amrmp_exps}
\input{tex/conclusion}

\bibliographystyle{IEEEtran}
\bibliography{references}

\end{document}

%% file: figures/fig_teaser.tex
\begin{strip}
\vspace{-2.0cm}
\begin{center}
  \includegraphics[height=2.3cm]{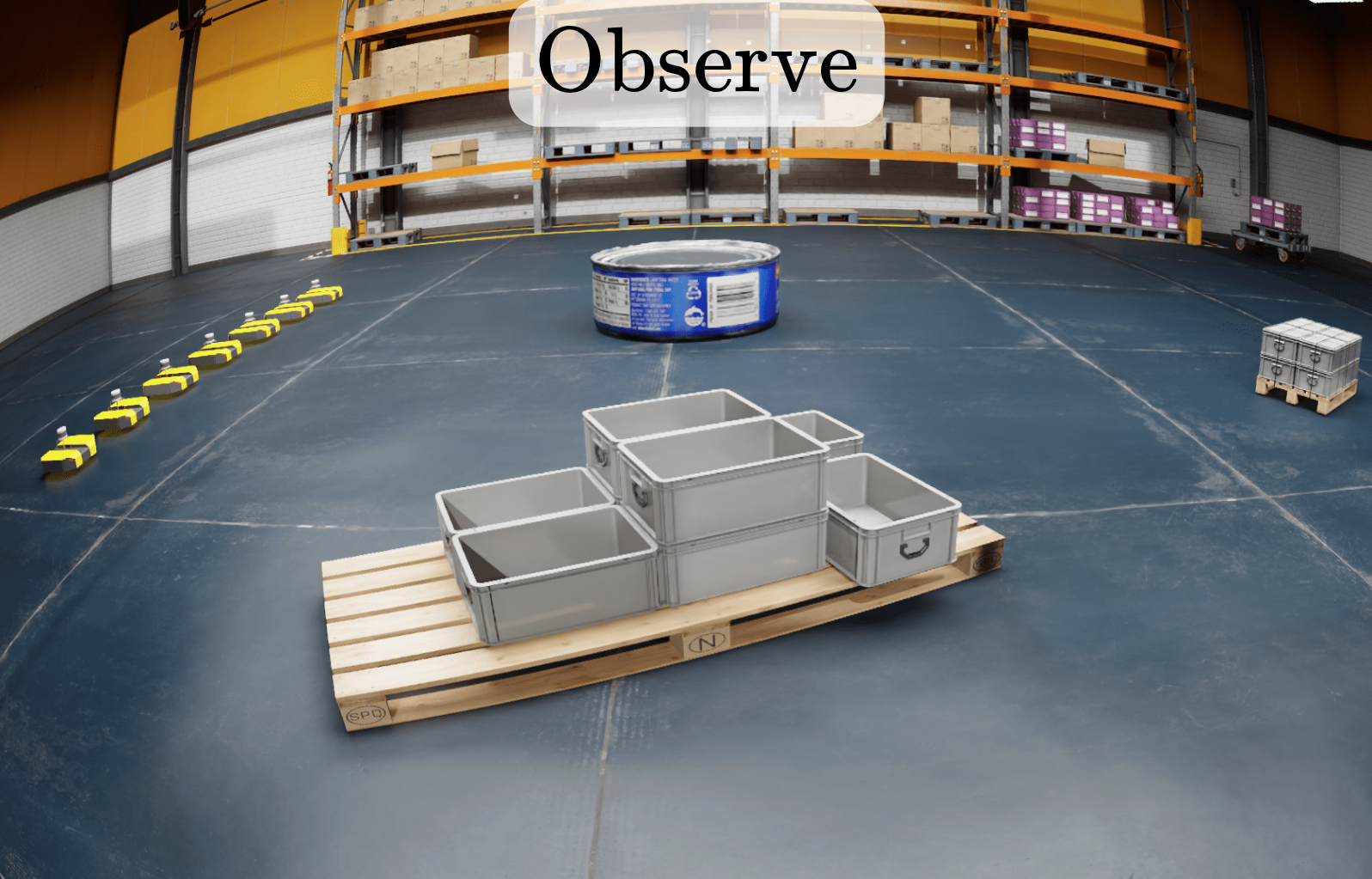}  
  \includegraphics[height=2.3cm]{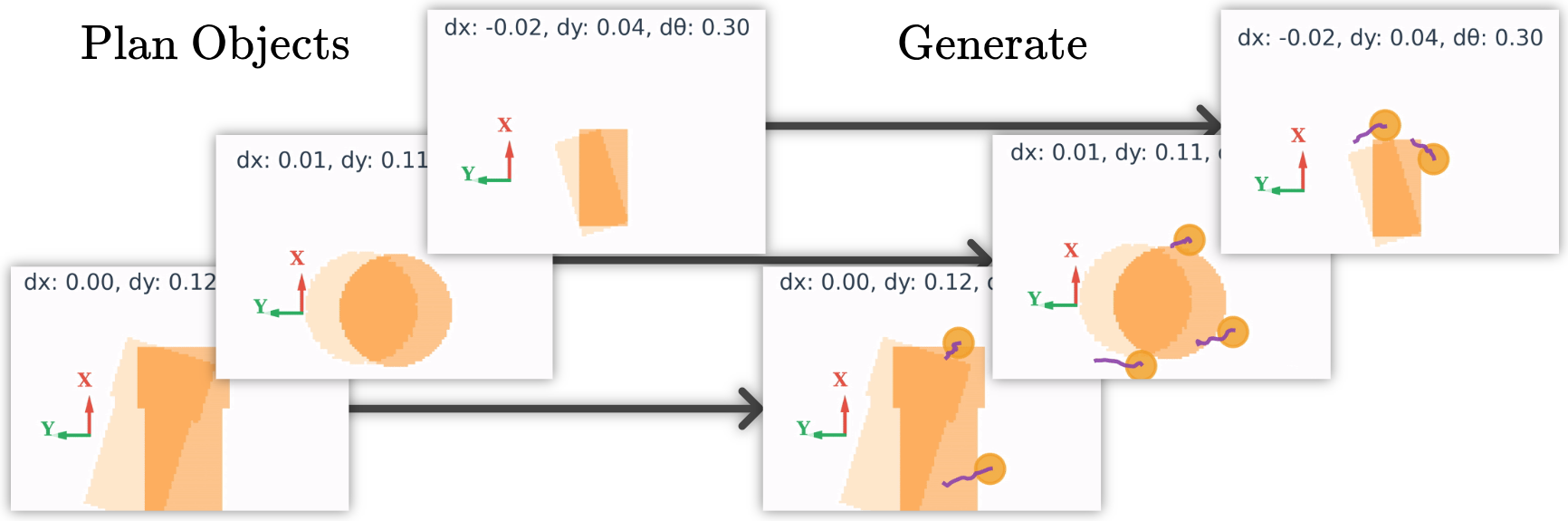} 
  \includegraphics[height=2.3cm]{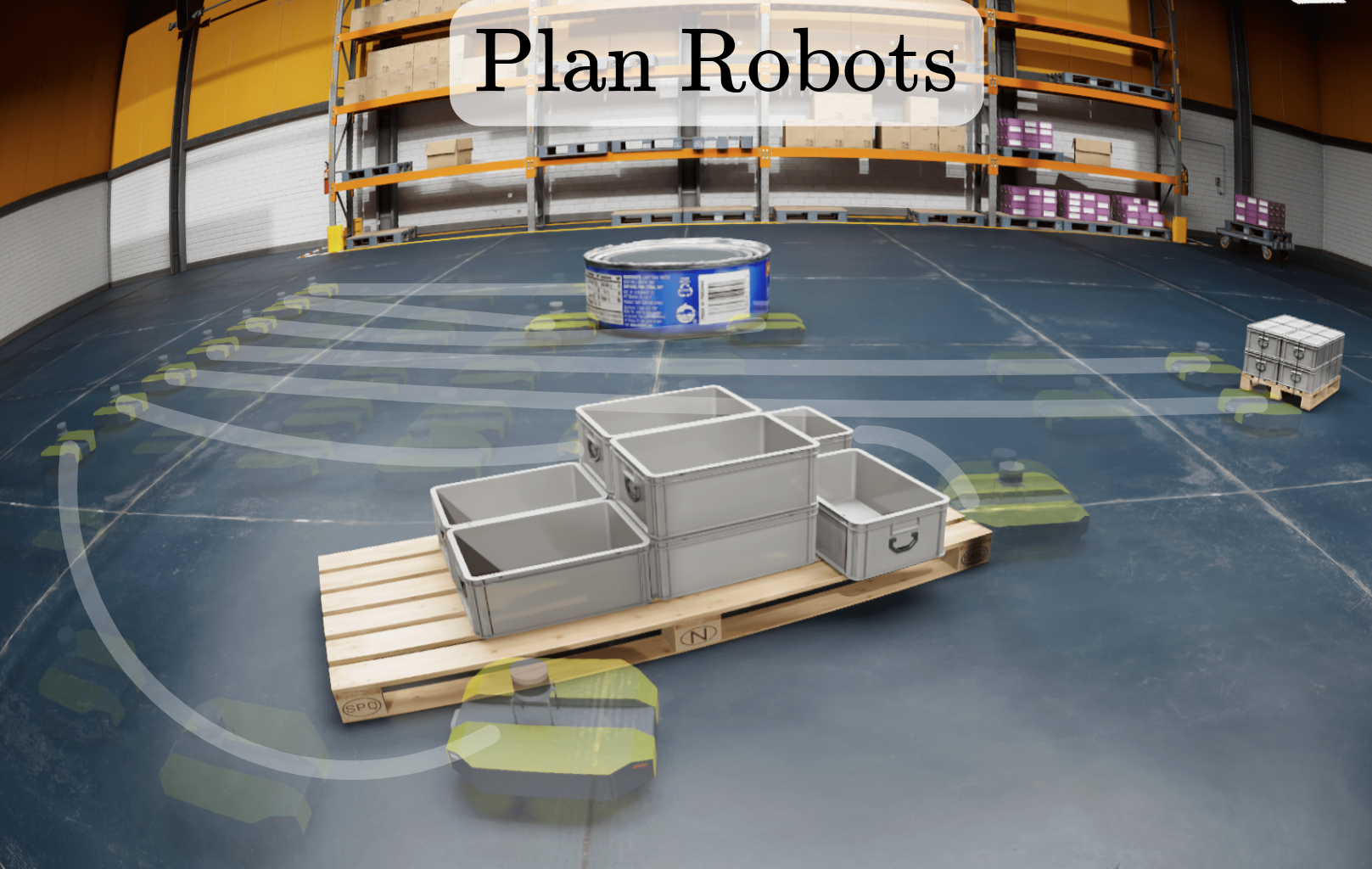}  
  \includegraphics[height=2.3cm]{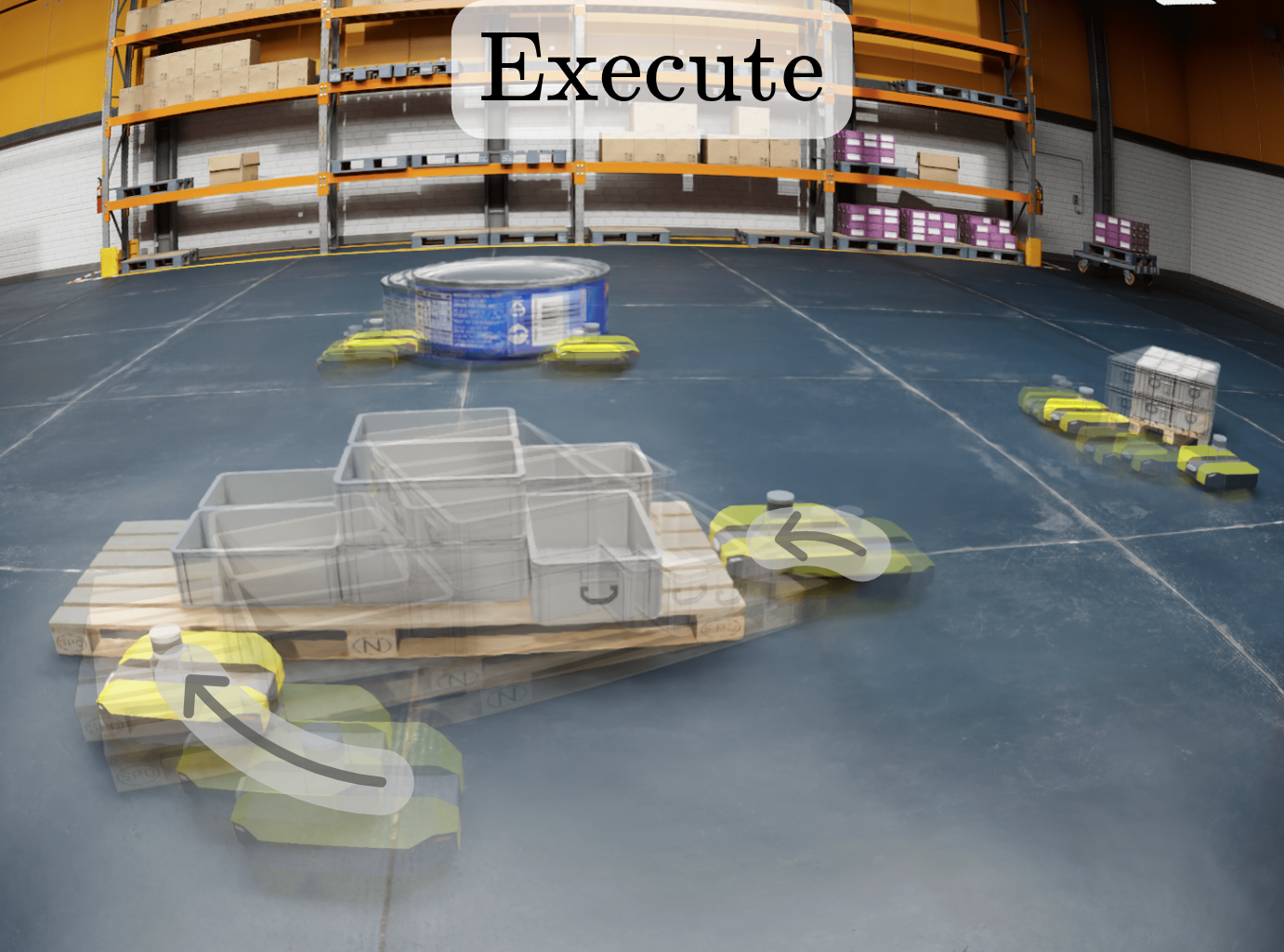}
  \captionof{figure}{The Generative Collaboration (\gco{}) framework learns components that are hard to model and plans those that are easy.
  From left to right: one iteration of \gco{}, coordinating seven robots collaboratively manipulating three large objects in a shared space.
  Given image observations, it proposes motions for all objects, jointly generates contact points and manipulation trajectories with flow matching co-generation, and plans multi-robot paths to convey the team to manipulation sites.}
  \label{fig:teaser}
\end{center}
\vspace{-0.6cm}
\end{strip}

%% file: tex/introduction.tex
\section{Introduction}
\label{sec:intro}
% The future holds the opportunity to have robots integrated into many facets of life, from the home to the warehouse. As systems improve and scale, it is inevitable for teams of robots to benefit from working together.
% To the best of out knowledge, this work is the first to employ flow matching co-generation in robotics.
Multi-robot collaboration is a central ambition in robotics, promising systems of robots that can accomplish tasks beyond the capabilities of an individual. A particularly compelling example is multi-robot manipulation, where robots must physically coordinate to move and position objects in their environment. Applications range from the home, where teams of robots tidy living spaces, to warehouse logistics, where robots transport and align inventory.

Multi-robot manipulation poses unique challenges. Unlike purely navigational coordination, robots must reason about their own motions, the coupled dynamics of objects under contact, and how to assign and reassign interactions across multiple robots and objects. This coupling makes the problem complex: the space of possible robot–object assignments, contact formations, and coordinated motions grows quickly with the number of robots and objects.

Prior methods approach this challenge from two extremes. \newline Learning-based approaches aim to capture the entire process from data, which can model complex dynamics but often sacrifices generalization and scalability \cite{feng2025mapush}. Planning-based approaches \cite{tang2024KG-HS} leverage structure to compute coordinated trajectories, but rely heavily on perfect models and can be brittle when deployed in the real world. In contrast, we argue for a middle ground: planning should be used wherever structure allows, and learning should be applied where models are uncertain or intractable to design by hand.

In this paper, we present a unified framework for collaborative manipulation that embodies this principle, with three contributions:
(i) {The Generative Collaboration framework} (\gco) for scalable multi-robot, multi-object collaborative manipulation. 
(ii) {A flow matching co-generation interaction model} that proposes feasible and diverse contact and motion strategies for objects observed through images alone. To the best of our knowledge, this is the first use of discrete-continuous flow matching co-generation in robotics.
% \footnote{Flow matching co-generation remains largely unexplored in robotics. The closest work \cite{hoeg2025HDP} couples discrete task sequencing with continuous motion using hybrid diffusion. Our approach differs by directly co-generating discrete \emph{contact points} \(\contactset{}\) and continuous \textit{trajectories} \(\robtrajset{}\) within a unified flow matching framework, enabling scalable collaborative manipulation.}
And 
(iii) {a new multi-robot motion planner} (\gspi{}) that computes safe, coordinated trajectories for conveying multiple entities to goal configurations.
We show that the \gco{} framework, and individually the \gspi{} algorithm, both effectively solve more complex and larger scale problems than existing methods to manipulation and multi-robot motion planning.

% Our approach enables teams of robots to manipulate multiple objects through non-prehensile interactions, while dynamically adapting to uncertainty via interleaved planning and execution. We evaluate our framework in simulated environments with varying numbers of robots, objects, and obstacles. Results demonstrate that coupling planning with interaction generation yields solutions that are more scalable, robust, and efficient than prior methods that attempt to learn the entire task end-to-end.

%% file: tex/problem.tex
\section{Problem Formulation}

% In this work, we explore multi-robot collaboration through the lens of collaborative manipulation.
% Consider a set $\robotset:=\{\robot{1}, \robot{2},\cdots \robot{N}\}$ of $N$ robots sharing a workspace $\mathcal{W} \subseteq \mathbb{R}^2$. We abstract away robot embodiments and represent them as disks with a fixed radius. Let the configuration space of each robot $\robot{i}$ be $\robcfg{i} \in \robcfgspace{i} \subseteq \mathbb{R}^2$.  A trajectory with horizon $H$ for robot $\robot{i}$ is a sequence of $H$ timed configurations and is denoted $\robtraj{i}{}$.  
% The world includes $M$ movable objects $\objset:= \{\obj{1}, \cdots \obj{M}\}$. An object configuration for $\obj{i}$ is denoted as  $\objcfg{i} \in \objcfgspace{i} \subseteq SE(2)$. The world may also have static obstacles.

% Given a set of $M$ object goal configurations $\objgoalcfg{i}$, the task at hand is to convey objects to goal configurations. In this work, we treat objcts as anonymous, and do not care about which object occupies which goal. This can be seen as laying bricks to build a structure. It need not matter which brick occupies which space. Objects can be moved by the robots via non-prehensile manipulation (e.g., by pushing). Thus, our task is to compute a set of trajectories $\robtrajset := \{\robtraj{i}{}\}_{i=1}^{N}$ whose execution would move objects to occupy all goals. Importantly, this does not mean that execution must be done in an open-loop manner. Our algorithm interleaves planning and execution to deal with execution uncertainty.

We study the problem of multi-robot collaborative manipulation, where a team of robots jointly transports multiple objects to designated goal configurations through non-prehensile interactions (i.e., pushing).

Let $\robotset = \{\robot{1}, \robot{2}, \ldots, \robot{N}\}$ denote the set of $N$ holonomic \textit{robots} operating in a shared workspace $\mathcal{W} \subseteq \mathbb{R}^2$, modeled as disks of fixed radius $r > 0$. The configuration of robot $\robot{i}$ is $\robcfg{i} \in \robcfgspace{i} \subseteq \mathbb{R}^2$, and a trajectory of horizon $H$ is defined as a sequence of timed configurations and termed $\robtraj{i}{}$, with $\robtraj{i}{t}$ denoting the robot configuration at time $t$.
% \in [0, H]$.

The environment also contains a set of $M$ movable \textit{objects}, $\objset = \{\obj{1}, \ldots, \obj{M}\}$, where a configuration of $\obj{j}$ is $\objcfg{j} \in \objcfgspace{j} \subseteq SE(2)$. The workspace may additionally contain static \textit{obstacles} that cannot be intersected by robots or objects.

A multi-object manipulation task specifies $M$ goal configurations $\objgoalcfg{1}, \ldots, \objgoalcfg{M}$ for objects. We adopt an anonymous assignment between objects and goals: any object can occupy any goal, and a problem is solved once all goals are filled. This naturally captures applications like warehouse distribution, where packages can be routed to any available shipping station within a set of acceptable options.

The task objective is to compute a set of coordinated robot trajectories
$\robtrajset{} = \{\robtraj{1}{}, \cdots , \robtraj{N}{}\}$
such that, when executed, the resulting coupled robot–object dynamics cause all objects in $\objset$ to reach the set of goal configurations $\{\objgoalcfg{i}\}_{i=1}^M$. Importantly, the trajectories need not remain fixed throughout execution. To cope with model mismatch and real-world uncertainty, our framework interleaves planning with execution, allowing dynamic replanning and reassignment of robot–object interactions as needed.

%% file: tex/related.tex
\section{Background}
\label{sec:related}

We now situate our work within the broader literature. We survey related work in Sec. \ref{sec:background_multi_robot_manip} and provide relevant background on flow matching for generative modeling (\ref{sec:background_fm}) and on anonymous multi-robot motion planning (\ref{sec:background_amrmp}).

\subsection{Related Work: Multi-Robot Manipulation}
\label{sec:background_multi_robot_manip}

Much of the research on multi-robot coordination has centered on geometric collision avoidance, where robots are tasked with reaching goal positions while avoiding one another and obstacles, mostly in simplified 2D grid environments \cite{sharon2015conflict, li2021eecbs, okumura2019pibt}, with some applied to robotic arms and drones \cite{shaoul2024accelerating, shaoul2024gencbs, honig2018trajectory}. Far less attention has been given to collaboration through interaction, where robots must not only avoid each other but also physically engage with objects.

Early approaches addressed only parts of the problem or relied on strong assumptions, avoiding the combinatorial and physical challenges of jointly reasoning about assignments, motions, and contact dynamics. For instance, Hartmann \textit{et al.}~\cite{Hartmann_rearrange}, addressed the task-and-motion planning (TAMP \cite{tamp}) formulation of multi-robot rearrangement, scaling impressively with objects and robots, but disregarded robot-object physical interactions. Tang \textit{et al.}~\cite{tang2024KG-HS} considered long-horizon planning for physical manipulation, but restricted attention to a single object whose exact shape and physical parameters are known, enabling geometric analysis.

Recent work has sought to relax these assumptions by learning all aspects of collaborative manipulation directly. Feng \textit{et al.}~\cite{feng2025mapush}, along with related efforts \cite{nachum2019multi, xiong2024mqe}, pursued hierarchical reinforcement learning (HRL), in which low-level controllers handled locomotion while higher-level policies determined how robots engage with objects. Although in principle such approaches could scale, in practice they have only been demonstrated on drastically simplified problems—typically no more than two or three robots, a single object, and environments with few or no obstacles. Even within these restricted settings, HRL policies often struggle with large transformations and obstacle avoidance. By contrast, our work introduces algorithms that scale with the number of robots and objects, even in cluttered spaces where robots may be outnumbered by objects or vice versa.

In this work, we depart from approaches that either plan under strong modeling assumptions or learn from scratch in restricted settings, and instead leverage structure where possible while learning only what is hard to model. We treat collision-free navigation and task assignment as structured subproblems amenable to planning, and learn contact strategies (i.e., how robots engage objects and coordinate their influence) from data. This perspective is supported by the recent success of diffusion and flow matching methods for learning local skills \cite{chi2023diffusionpolicy, black2024pizero, carvalho2023mpd} and by the ability of planning algorithms to globally coordinate large robot teams \cite{okumura2023tswap, okumura2019pibt}.

\subsection{Background: Flow Matching Co-Generation}
\label{sec:background_fm}
Flow matching, a flexible method for generative modeling, has recently found success in various domains, including imitation learning for robotics \cite{black2024pizero}, image generation \cite{lipman2022flow}, and protein design \cite{campbell2024multiflow}. In this work, we leverage flow matching models to determine where robots should create contact with objects, how many contact points to create, and how to move once contact is established. Let us begin with background on three flavors of flow matching: continuous, discrete, and continuous-discrete co-generation.

\textbf{Continuous Flow Matching} casts generation as learning a time-dependent velocity field $u_t$ that transports a base distribution $p_0$ into the data distribution $p_1$ \cite{lipman2022flow}. For an interpolated sample ${^{t}\!X} := ({^{1}\!X} - {^{0}\!X})t + {^{0}\!X}$
between a noisy sample ${^{0}\!X} \sim p_0$ and a data sample ${^{1}\!X} \sim p_1$, the flow matching objective is to learn an estimated velocity field $u_t^\theta$ that matches the sample velocity. 
$$\theta^* = \arg\min_{\theta} \mathbb{E}_{{^{0}\!X}\sim p_0, {^{1}\!X}\sim p_1, t \in [0,1]} \Vert ({^{1}\!X} - {^{0}\!X}) - u^\theta_t({^{t}\!X}) \Vert^2.$$

\noindent Generating new samples $\hat {^{1}\!X}$ is done by integrating the ODE
\begin{equation}
\label{eq:ode_integration}
\frac{d}{dt} {^{t}\!X} = u_t^\theta({^{t}\!X}), 
\ {^{0}\!X} \sim p_0, \quad {^{1}\!X} = {^{0}\!X} + \int_{0}^{1}u^\theta_t({^{t}\!X}) dt.
\end{equation}

While flow matching is powerful for continuous data (e.g., trajectory waypoints), categorical domains where samples are sequences of values from a fixed set (e.g., choosing contact points from a finite set of image pixels or text generation~\cite{gat2024discreteflowmatching}) benefit from a different treatment.

\textbf{Discrete Flow Matching} \cite{gat2024discreteflowmatching} is a framework for generating sequences $K$ of discrete samples when elementes can only take on values from a discrete, finite, state space $\contactspace$. Let a sample from the space of discrete sequences of length $B\in\mathbb{Z}^{>0}$ elements be $K := \{K^{(1)}, \cdots K^{(B)} \mid K^{(b)} \in \contactspace\}$.
Given ``noise'' and data sequences ${^{0}\!K} \sim p_{\text{d}, 0}$ and ${^{1}\!K} \sim p_{\text{d}, 1}$, an interpolated sample ${^{t}\!K}$ is obtained by mixing
$$
{^{t}\!K^{(b)}} \sim \text{Cat}\big((1-t)\delta_{{^{0}\!K^{(b)}}} + t \delta_{{^{1}\!K^{(b)}}}\big),
$$
where $\delta$ is the Dirac delta function and Cat is the categorical distribution. Simply put, value $K^{(b)} \in {^{t}\!K}$ takes on the base value ${^{0}\!K}^{(b)}$ with probability $(1-t)$ or the data value ${^{1}\!K}^{(b)}$ with probability $t \in [0, 1]$.

The discrete flow matching objective is to predict $\delta_{{^{1}\!K}}$ directly from an interpolated ${^{t}\!K}$. This prediction, which is sampled from the generated distribution $u^\phi_{\text{d}, t}({^{t}\!K})$ named the discrete velocity, is supervised with a cross-entropy (CE) loss
% over a factorized distribution
\vspace{-2mm}
\[
\vspace{-1.5mm}
\phi^* = \arg\min_{\phi}\,
\mathbb{E}_{{^{0}\!K}, {^{1}\!K}, t} 
% _{{^{0}\!K} \sim p_{\text{d}, 0},\, {^{1}\!K} \sim p_{\text{d}, 1},\, t \in [0,1]} 
\left[
\sum_{b=1}^{B} 
\mathrm{CE}\!\left(u^\phi_{\text{d}, t}({^{t}\!K}^{(b)}), \, \delta_{{^{1}\!K}^{(b)}}\right)
\right]
\]
with ${{^{0}\!K} \sim p_{\text{d}, 0},\, {^{1}\!K} \sim p_{\text{d}, 1},\, t \in [0,1]}$.
\noindent Generating new discrete samples $\hat {^{1}\!K}$ is done by simulating a categorical Markov chain, where each element $K^{(b)}$ transitions according to the predicted discrete velocity $u^\phi_{\text{d},t}$ over $t \in [0,1]$,
\begin{equation}
\label{eq:discrete_fm_evolution}
    {^{0}\!K} \sim p_{\text{d}, 0}, \quad
    ^{{t+\Delta t}\!}K^{(b)} \sim \text{Cat}\big( u^\phi_{\text{d}, t}({^{t}\!K}^{(b)}) \big)
\end{equation}
\noindent The base distribution $p_{\text{d}, 0}$ we adopt in this work is the ``mask'' distribution where all elements are set to a single value \texttt{[M]}.

% \textbf{Flow-Matching Co-Generation} \cite{campbell2024multiflow} attempts to jointly generate tightly coupled discrete and continuous samples. They do so by generating a discrete velocity and a continuous velocity together, and integrating them at the same time to generate new samples.

% Finding early success in RNA and protein sequence and structure generation \cite{rubin2025ribogen, campbell2024multiflow}, we see promise in applying these methods to problems in robotics where discrete (e.g., task assignment) and continuous (e.g., motion planning) values are often co-dependent \cite{garrett2020pddlstream, hoeg2025HDP}. 
% % \citet{hoeg2025HDP} recently explored discrete-continuous diffusion; to our knowledge, flow-matching co-generation remains unexplored in robotics.

\textbf{Flow Matching Co-Generation} \cite{campbell2024multiflow} attempts to jointly generate tightly coupled discrete and continuous samples. Let a joint state be $({^{t}\!X}, {^{t}\!K})$ where ${^{t}\!K}$ denotes discrete valued sequence and ${^{t}\!X}$ denotes continuous valued samples. A co-generative flow matching model learns a pair of fields
\vspace{-1mm}
\[
\vspace{-1mm}
u^\phi_{\text{d}, t}({^{t}\!X}, {^{t}\!K}), \quad u^\theta_{\text{c}, t}({^{t}\!X}, {^{t}\!K}),
\]
where $u^\theta_{\text{c}, t}$ is a continuous velocity and $u^\phi_{\text{d}, t}$ is a discrete velocity, both conditioned on the discrete and continuous variables. The training objective is the sum of continuous and discrete flow matching losses, and generating new joint samples $(\hat {^{1}\!X}, \hat {^{1}\!K})$ is done by simultaneously integrating the continuous ODE (Eq. \ref{eq:ode_integration})
and evolving the discrete state ${^{t}\!K}$ via a categorical Markov chain with transition probabilities given by $u^\phi_{\text{d}, t}({^{t}\!X}, {^{t}\!K})$ (Eq. \ref{eq:discrete_fm_evolution}), both starting from $({^{0}\!X}, {^{0}\!K}) \sim (p_0, \delta_{\texttt{[M]}})$. In this work, we show that flow matching co-generation can be an effective choice for learning contact points and manipulation trajectories.

\subsection{Background: AMRMP}
\label{sec:background_amrmp}

Anonymous multi-robot motion planning (AMRMP) is a foundational problem in robotics: given $N$ robots with start configurations $\robstartcfgset \in \robcfgspace{1} \times \robcfgspace{2} \times \cdots \times \robcfgspace{N}$ and a set of interchangeable goals $\robgoalcfgset$, the task is to compute safe trajectories $\robtrajset{}$ such that the collection of terminal states coincides with $\robgoalcfgset$.  

Most prior work on AMRMP studied the graph-based abstraction known as anonymous multi-agent path finding (AMAPF), where robots are modeled as points moving along vertices and edges \cite{yu2013amapfflow, honig2018cbsta, okumura2023tswap, tang2024itaecbs}. Within this setting, most methods focus on optimal or bounded sub-optimal path finding and goal assignments, often building on Conflict-Based Search (CBS) \cite{sharon2015conflict} to leverage its strong theoretical guarantees. Other approaches emphasize scalability over optimality. TSWAP \cite{okumura2023tswap}, for instance, is a highly efficient rule-based algorithm that detects potential deadlocks and resolves them by swapping goals between robots.  

Encouraged by the success of algorithms in these discrete point-robot domains, recent works have extended MAPF (i.e., the non-anonymous counterpart to AMAPF) to continuous settings by replacing grid edges with \emph{motion primitives}, short kinematically feasible transitions between configurations \cite{cohen2010search, shaoul2024accelerating, shaoul2024gencbs}. Unfortunately, similarly adapting AMAPF algorithms is not trivial: CBS-based methods lose much of their efficiency without grid-specific enhancements, and highly efficient rule-based methods like TSWAP can deadlock when applied to non-point robots and motion primitives.

In this work, we introduce \gspi{} (``Giuseppe''), a new AMRMP algorithm that combines the efficiency of rule-based MAPF methods with the deadlock resolution strategies inspired by TSWAP and C-UNAV~\cite{dergachev2024dec-unav}, adapted to motion-primitive domains without introducing deadlocks or livelocks in practice. \gspi{} forms a central building block of our generative collaboration framework, and with this background in place we now turn to its introduction.
\vspace{-2mm}

%% file: tex/method.tex
\begin{algorithm}[h!]
\vspace{-0.1cm}
\caption{\gco{}: Generative Collaboration for Multi-Robot Manipulation}
\label{alg:gco}
\footnotesize
\begin{algorithmic}[1]
\Require World state $\mathcal{X}_w$, object set $\objset$, goal set $\{\objgoalcfg{i}\}_{i=1}^M$, robot set $\robotset$, policy $\pi_\theta$, planner $\gspi$
\While{not all objects in goal set}
  \State $\mathcal{X}_w,\; \{\image{j}\}_{j=1}^M \gets \textsc{Observe}()$  \label{line:observe}
  \State $\{\objtraj{j}\}_{j=1}^M \gets \textsc{Plan}(\gspi, \objset, \mathcal{X}_w)$ \label{line:objplan}
  \State $\{T^j\}_{j=1}^M \gets \textsc{ExtractTransforms}(\{\objtraj{j}\})$ \label{line:transform} 
  \State $\{B^j\}_{j=1}^M \gets \textsc{BudgetAllocation}(\robotset, \objset)$ \label{line:budget}
  \State $\{\contactset{j}, \robtrajset{j}_{\text{manip}}\}_{j=1}^M \gets \pi_\theta(\{\image{j}, T^j, B^j\}_{j=1}^M)$ \label{line:policy}
  \State $\robtrajset{}_{\text{coop}} \gets \textsc{Plan}(\gspi, \robotset, \{\contactset{j}\})$ \label{line:robotplan}
  \State \textsc{Execute}($\robtrajset{}_{\text{coop}} \oplus 
 \cup\{\robtrajset{j}_{\text{manip}}\}_{j=1}^M$) \label{line:execute} 
\EndWhile
\end{algorithmic}
\end{algorithm}
\vspace{-0.4cm}
\section{Generative Collaboration}
\label{sec:method}
% Despite substantial advances in multi-robot planning and learning, existing approaches fall short of addressing collaboration at scale in a physically grounded way. To this end, w
We introduce \gco{}, a framework for interleaving learning and planning in collaborative multi-robot systems. The core principle of \gco{} is simple: \textbf{plan what can be modeled well, and learn what cannot.}

In the context of non-prehensile manipulation, we identify two distinct operation modes that must be addressed by a team of robots: they must generate manipulation motions that define how robots interact with objects to induce desired transformations, and they must generate cooperative motions that bring robots into the appropriate configurations to realize those interactions. Manipulation motions depend on objects’ physical dynamics as well as stochasticity in execution—making them natural candidates for learning. In contrast, cooperative motions involve the geometric feasibility of moving multiple entities through clutter, a domain amenable to planning. \gco{} unifies these two modes in a closed-loop framework, interleaving generation of data-driven manipulation motions and planning of cooperative trajectories to realize them (Alg.~\ref{alg:gco}). 

Each iteration begins with perception (line~\ref{line:observe}). 
Objects $\obj{j}$ are observed via images $\image{j}$, which are $w\times w$ occupancy matrices, and the current world state $\mathcal{X}_w$ is updated with the states of robots and obstacles.
Next, the planner $\gspi$ is invoked at the object-level to generate short-horizon trajectories $\objtraj{j}$ for each object (line~\ref{line:objplan}), from which short transformation subgoals $T^j$ are extracted (line~\ref{line:transform}). Robots are then budgeted\footnote{ In this work, we allocate robots as evenly as possible across all objects that need to be manipulated, with idle robots sent to wait at safe locations. When objects outnumber robots, robots are assigned by proximity. }
across all objects with non-zero $T^j$, each receiving $B^j$ robots (line~\ref{line:budget}). Finally, the learned manipulation policy, termed $\pi_\theta$, co-generates object-centric contact points $\contactset{j}$ and manipulation trajectories $\robtrajset{j}_{\text{manip}}$ conditioned on images, transforms, and robot budgets (line~\ref{line:policy}).
% To realize these motions, \gspi{} is invoked a second time at the robot-level to compute cooperative trajectories $\robtrajset{}_{\text{coop}}$ that convey robots to the assigned contact points (line~\ref{line:robotplan}). Finally, the free-space cooperative trajectories for each robot are concatenated with the manipulation trajectories stemming from the contact points they terminate at and executed (line~\ref{line:execute}), advancing the system toward the goal set. This loop continues until all object-goals are achieved. 
To realize these motions, \gspi{} is called a second time at the robot level to compute cooperative trajectories $\robtrajset{}_{\text{coop}}$, which bring robots to contact points (line~\ref{line:robotplan}). Once computed, these free-space cooperative trajectories are concatenated with the corresponding manipulation trajectories that begin at the reached contact points and are then executed (line~\ref{line:execute}). This process advances the system toward the object-goal set, and the loop repeats until all object-goals have been achieved.

% The procedure is summarized in Alg.~\ref{alg:gco}. Each iteration begins with perception (line~\ref{line:observe}), followed by short-horizon object-level planning (line~\ref{line:objplan}) and subgoal extraction (line~\ref{line:transform}). Robots are then budgeted across objects (line~\ref{line:budget}) before the learned policy $\pi_\theta$ co-generates contact points and manipulation trajectories (line~\ref{line:policy}). Generation is accelerated using batch-processing and can be done at $\sim 50\; \mathrm{Hz}$. Finally, \gspi{} is invoked again at the robot level to plan cooperative motions (line~\ref{line:robotplan}), which are concatenated with manipulation trajectories for execution (line~\ref{line:execute}). This cycle repeats until all object-goals are achieved.

We detail \gco{}’s manipulation module in Sec.~\ref{sec:method_fm}, describe \gspi{} in Sec.~\ref{sec:method_gspi}, and present experiments comparing \gco{} and \gspi{} to existing baselines in Sec.~\ref{sec:experiments}.

% ================================================
% Flow matching for manipulation
% ================================================
\subsection{Flow Matching Co-Generation for Manipulation}
\label{sec:method_fm}
Our objective is to endow teams of robots with the ability to generate object-level manipulation motions that achieve target transformations $T^j \in SE(2)$. Central to this problem is deciding both \emph{where} robots should make contact with objects and \emph{how} they should move once contact is established. We pursue a flow matching approach and introduce three instantiations: \gcoct{}, which treats the task as a monolithic synthesis problem by directly generating continuous trajectories without explicit contact modeling; \gcocc{}, which co-generates contact points and trajectories jointly in continuous space, parameterizing contacts in the object frame and trajectories as short waypoint sequences; and \gcodc{}, which anchors contact points in the perceptual space through discrete selection while generating trajectories continuously. All three methods are effective and consistently outperform baselines, while the co-generation variants provide added stability, with the discrete–continuous formulation proving the most reliable in challenging tasks (Sec.~\ref{sec:experiments}).

\subsubsection{{Direct Continuous Trajectory Generation (\gcoct{})}}  
In \gcoct{}, multi-robot manipulation is posed as generating robot trajectories, treating their first state as a contact point:  
\[
% \vspace{-2mm}
\robtrajset{j}_{\text{manip}} \in \mathbb{R}^{B_{\max} \times H \times 2}.
\]  
I.e., $\robtrajset{j}_{\text{manip}}$ contains $B_{\max}$ trajectories (a maximum budget), each of length $H$, whose values are unconstrained. Executing $\robtrajset{j}_{\text{manip}}$ should transform $\obj{j}$ by $T^j$.
To generate, noise $^0\robtrajset{j}_{\text{manip}} \sim \mathcal{N}(\mathbf{0}, \mathbf{I})$ is evolved into structured trajectories via integration of a learned conditional velocity field $u_t^\theta$:  
\[
\robtrajset{j}_{\text{manip}} = \int_0^1 u_t^\theta(^t\robtrajset{j}_{\text{manip}} \mid \image{j}, T^j, B^j) dt.
\]  

While conceptually simple, generating unconstrained sequences of waypoints for all robots can be difficult to learn.
%  this approach requires reasoning about a high-dimensional space; 

\subsubsection{{Continuous--Continuous Co-Generation (\gcocc{})}}  
\gcocc{} decomposes generation into two continuous components: contact points in the local object frame, $\contactset{j}$, and manipulation trajectories rooted at the origin, $\robtrajset{j}_{\text{manip}}$,
\[
\contactset{j} = \{\contact{j, b}{} \in \mathbb{R}^2\}_{b=1}^{B_{\max{}}}, \qquad
\robtrajset{j}_{\text{manip}} = \{\robtraj{j,b}{} \in \mathbb{R}^{H \times 2}\}_{b=1}^{B_{\max{}}}.
\]
For execution, $\robtraj{j,b}{}$ are translated to begin at points $\contact{j,b}{}$.

A flow matching model learns two velocity fields, $u^\theta_t$ and $u^\phi_t$ for contacts and trajectories, and jointly evolves $(\contactset{j}, \robtrajset{j}_{\text{manip}})$ from noisy seeds $(^0\contactset{j}, ^0\robtrajset{j}_{\text{manip}})$. In practice, we couple Euler integration for both continuous components:
% to obtain $^{1\!}\contactset{j}, ^{1\!}\robtrajset{j}_{\text{manip}}$:
\begin{align*}
^{t+\Delta t}\contactset{j} &=\; ^{t\!}\contactset{j} + u_t^\theta(^t\contactset{j} \mid\; ^{t\!}\robtrajset{j}_{\text{manip}}, \image{j}, T^j, B^j) \Delta t \\
^{t+\Delta t} \robtrajset{j}_{\text{manip}} &=\; ^{t\!}\robtrajset{j}_{\text{manip}} + u_t^\phi( ^{t\!}\robtrajset{j}_{\text{manip}} \mid\;
^t\contactset{j}, \image{j}, T^j, B^j) \Delta t.
\end{align*}

This decomposition reduces complexity by explicitly modeling structure---where to create contact, and how to move---but still requires continuous contact generation, which unnecessarily reasons over a space larger than that perceived.  

\subsubsection{{Discrete--Continuous Co-Generation (\gcodc{})}} 
Our key contribution is the discrete--continuous formulation \gcodc{}, which couples the two components in {different representational spaces}. We introduce {discrete contact point generation}, where contacts are selected directly from the finite set of pixels in the observed image. We write a state $\contactset{j}$, holding information about $B_{\max{}}$ contact points for an object $\obj{j}{}$, as a sequence of $2B_{\max{}}$ discrete tokens $\{\tok{j}{i}\}_{i=1}^{2B_{\max{}}}$, with each consecutive pair being equivalent to an $(x,y)$ pixel coordinate in the image frame. Formally:

\vspace{-2mm}
{\scriptsize
\[
    \vspace{-2mm}
    \contactset{j} := \bigg\{\underbrace{\tok{j}{1}, \tok{j}{2}}_{\contact{j}{1}}, \dots,
    \underbrace{\tok{j}{2 B_{\max{}}-1}, \tok{j}{2 B_{\max{}}}}_{\contact{j,B_{\max{}}}{}} 
    \;\bigg\vert\; 
    \tok{j}{i} \in \contactspace{}{} = \{1,\dots w\} \cup \{\texttt{[M]}\}
\bigg\}.
\]
}

Notationally, we write $(\contactset{j})^{(i)}$ to denote the $i^\text{th}$ token in the state. This formulation anchors the contact point space to perceptual evidence while the set of manipulation motions remains flexible in $\mathbb{R}^{B_{\max{}} \times H \times 2}$. Like before, each trajectory is rooted at the origin and later shifted to its contact point.

To generate new contact points and manipulation trajectories $\contactset{j}, \robtrajset{j}_{\text{manip}}$, we couple Euler integration (continuous component) with Markov chain transitions  (discrete component):
\begin{align*}
(^{t+\Delta t}\contactset{j})^{(i)} &\sim \text{Cat}\left( \left(u_{\text{d},t}^\theta(^t\contactset{j} \mid\;
^{t\!}\robtrajset{j}_{\text{manip}}, \image{j}, T^j, B^j \right)^{(i)} \right) \\
^{t+\Delta t} \robtrajset{j}_{\text{manip}} &= {^{t}\!}\robtrajset{j}_{\text{manip}} + u_{\text{c},t}^\phi(^{t\!}\robtrajset{j}_{\text{manip}} \mid\;
^t\contactset{j}, \image{j}, T^j, B^j) \Delta t.
\end{align*}

% \noindent A flow-matching co-generative model simultaneously evolves the discrete and continuous variables:  
% \[
% \contactset{j}, \robtrajset{j} = f^\theta(^0\contactset{j}, ^0\robtrajset{j} \mid \image{j}, T^j, B^j),
% \]  
% where $^0\contactset{j}$ are masked pixel seeds and $^0\robtrajset{j}$ are noisy continuous trajectories. The discrete flow-matching component evolves pixel indices over time, while the continuous component integrates a velocity field in trajectory space. Fig.~\ref{fig:gco_model_diagram} illustrates this process.  

\subsubsection{{Variable Robot Budget}}  
\label{sec:budget}
In practice, only a limited number of robots may be available for manipulation. Training separate models for each team size would be inefficient. Instead, we train a single model that respects a robot budget $B$. Our model always produces $B_{\max{}}$ contacts and manipulation trajectories, and masks entries that are not needed. We use the special token \texttt{[M]} to denote unused outputs: in the discrete case, $\contact{j,b}{}=(\texttt{[M]}, \texttt{[M]})$; in the continuous case, contacts within $\varepsilon$ of the mask token's value are treated as masked, i.e., when $\Vert \contact{j,b}{} - (\texttt{[M]}, \texttt{[M]}) \Vert < \varepsilon$.  
In training, supervision penalizes generated budgets $\hat{B}^j$ exceeding the allocated budget $B^j$: $\mathcal{L}_\text{budget} = \sum_{j=1}^M \mathrm{ReLU}(\hat{B}^j - B^j)$.

\subsection{\gspi{}: Robot- and Object-Level AMRMP}
\label{sec:method_gspi}

% Once manipulation strategies are generated by $\pi_\theta$ (Sec.~\ref{sec:method_fm}), robots must be routed to contact points without collisions or deadlocks. 

In \gco{}, planned objects transformations are supplied to $\pi_\theta$ (Sec.~\ref{sec:method_fm}) for generating manipulation strategies, and robots plan motions to reach contact points.
Both planning stages falls under AMRMP: entities move in continuous space, are indistinguishable, and must reach a set of interchangeable goals even when operating in tight proximity to obstacles. Existing AMRMP planners are difficult to apply directly; while they handle open spaces well, they struggle when robots must approach surfaces or navigate narrow clearances (Fig.~\ref{fig:gspi_all_results})—a key requirement in our setting.

One might instead extend scalable AMAPF algorithms from point-robot grids to continuous motion-primitive domains, but na\"ive adaptations either impose coarse discretizations or introduce deadlocks (Sec.~\ref{sec:background_amrmp}). To address these challenges, we combine the efficiency of the grid-based method PIBT with goal-swapping ideas inpired by TSWAP and C-UNAV to develop \gspi{}, a new AMRMP algorithm that scales robustly in continuous, cluttered environments. Specifically, \gspi{} generalizes PIBT—a non-anonymous algorithm originally designed for point robots on regular grids—to the anonymous setting with circular robots and motion primitives. The remainder of this section details the design of \gspi{}, starting from PIBT and outlining how our proposed algorithm overcomes its limitations.

\input{tex/gspi_algorithm}
\subsubsection{{Priority Inheritance with Backtracking (PIBT)}}
% Before describing \gspi{}, we first revisit the logic of PIBT~\citep{okumura2019pibt}, on which our method builds. Alg.~\ref{alg:pigs-succinct} shows PIBT in black and the \gspi{} modifications in blue. 
PIBT~\cite{okumura2019pibt} is a multi-agent path-finding algorithm for point robots traversing edges and vertices on a shared graph. Shown in Alg. \ref{alg:pigs-succinct} in black, PIBT assigns each robot a dynamically changing priority (line~\ref{line:initial_priorities}) and operates in discrete timesteps. At each step, PIBT iterates through robots by priority and asks each to commit to an action (\textsc{TryMove} in line~\ref{line:try-move-loop}).
On a robot’s turn, it first sorts its available actions (transitions along edges incident to its current vertex) by a heuristic—typically the progress an action makes toward its goal (line~\ref{line:gen-mprims})—and prunes out any actions that collide with obstacles or higher-priority robots. From the remaining valid transitions, if its best option does not collide with any other robot’s position, it commits to that action (line \ref{line:no-conflicts}). Otherwise, it identifies the blocking robot and requests that it moves away (line \ref{line:try_move_permutation}). This is done by calling \textsc{TryMove} on the blocking robot, temporarily granting it the higher robot’s priority so it can ask lower-priority robots to make way for it. This recursive chain continues until all robots commit to collision-free actions, accommodating the request of the initiating robot. If a recursive call fails—meaning a robot cannot find a feasible move—the algorithm backtracks and tries the next-best action for the parent robot. If all options fail, the robot waits (line~\ref{line:wait-edge}). Finally, all committed motions are executed together, completing one timestep (line \ref{line:commit}).

\input{figures/fig_gco_manip_exps}
\subsubsection{{Limitations of PIBT}}
An important detail in PIBT is that priorities of robots en route to their goals gradually increase, while those of robots that reach their goals are reset to their lowest level (line~\ref{line:commit}). This rule gives active robots precedence over stationary ones. However, it can also create \emph{livelocks}: when goals and paths overlap, a robot at its goal may be pushed away by another. Once the displacing robot reaches its own goal, its priority drops in turn, allowing the first robot to push back. The two can continue alternating endlessly. This failure mode stems from PIBT fixed goal assignments.
In our manipulation framework, any robot can occupy any contact point. Capitalizing on this flexibility, we develop \textbf{G}oal \textbf{S}wapping with \textbf{P}riority \textbf{I}nheritance (\gspi{}): an AMRMP planner for non-point robots with motion primitives that shares PIBT’s lightweight structure without suffering from livelocks in practice. 

\subsubsection{{The \gspi{} Algorithm}}
At its core, \gspi{} follows the structure of PIBT and operates in iterations, where in each iteration robots commit to actions and execute them. In PIBT, each iteration consists solely of a \emph{move} stage (line~\ref{line:try-move-loop}), where robots select and execute actions. In \gspi{}, we add the \emph{swap} stage (line~\ref{line:try-swap-loop}), where robots may exchange goals.

The \emph{move stage} resembles that of PIBT but includes modifications that extend it to continuous domains with disk-shaped robots. In PIBT, 
% where robots are modeled as points, 
each candidate action corresponds to traversing a graph edge and can, at most, be blocked by a single other robot occupying the destination vertex. In contrast, \gspi{} operates with motion primitives and non-zero robot footprints. As a result, a candidate action may intersect several lower-priority robots if their configurations overlap with the primitive. 
To handle this, \gspi{} constructs the set $\mathcal{S}$ of all affected robots (line~\ref{line:lower-priority-stepped-on-set}) and asks them to yield to the higher-priority robot by creating a recursive \textsc{TryMove} call for each one, with later recursive calls treating earlier ones as higher-priority (line~\ref{line:try_move_permutation}). \gspi{} explores all orderings of affected robots in $\mathcal{S}$ (line~\ref{line:recursive-moves}); if one ordering succeeds, the motion is accepted, and otherwise, the system rolls back (line~\ref{line:rollback}) and tries the next candidate primitive. 
% This procedure yields coordinated motion in densely packed scenes.

In the \textit{swap stage}, \gspi{} iterates over robots by priority and evaluates whether exchanging goals and priorities between pairs would improve overall progress. A swap is proposed for a pair when two conditions hold (line~\ref{line:swap_if_conditional}): (i) the higher-priority robot is closer to the other’s goal than to its own, and (ii) the exchange does not increase the total team distance. If both are satisfied, the robots swap goals and priorities (lines~\ref{line:swap-goals} and~\ref{swap-priorities}).
The goal-swapping stage serves two key purposes. First, it relaxes fixed robot–goal assignments and promotes efficient routing in the \emph{anonymous} case, where any robot may occupy any goal. Second, in practice, it eliminates the livelocks that arise in PIBT’s fixed-goal formulation, where robots can repeatedly push one another without either remaining at a goal. Intuitively, livelocks occur in PIBT when a high-priority robot pushes a lower-priority robot off its goal and continues to push until reaching its own goal. In \gspi{}, such a situation triggers a goal swap: the higher-priority robot takes the closer goal, removing the need for the two robots to move head-to-head, and avoids a livelock. 
% Goal swaps are accompanied by priority swaps to maintain consistent goal–priority pairings, ensuring that, no matter which robot is working towards reaching a goal, it is always 
As we show in Sec. \ref{sec:exp_amrmp}, \gspi{} is a fast and scalable AMRMP algorithm. 

In our manipulation framework \gco{}, we use \gspi{} both to convey robots to their assigned contact points (Alg.~\ref{alg:gco}, line~\ref{line:robotplan}) and to compute transformations $T^j$ for each object $\obj{j}$ (Alg.~\ref{alg:gco}, lines \ref{line:objplan} and \ref{line:transform}). For the former, we run \gspi{} at the robot level until all robots occupy goals, and for the latter, we invoke \gspi{} at the object level for two iterations. 

%% file: tex/gspi_algorithm.tex
\vspace{-0.2cm}
\begin{algorithm}[h!]
\caption{\gspi{}: Goal Swapping with Priority Inheritance \newline \scriptsize{\textcolor{blue}{Blue lines are modifications to the PIBT algorithm.}}}
\label{alg:pigs-succinct}
\footnotesize
\begin{algorithmic}[1]
\Require \textcolor{blue}{Robot set $\robotset=\{\robot{1},\ldots,\robot{N}\}$ with configurations $\robcfg{i}\in\robcfgspace{i}$, 
anonymous goals $\robgoalcfgset$, motion-primitive generators $\mprimgen{i}(\robcfg{i})$, initial priority order $\prec$, current target map $g:\robotset\!\to\!\robgoalcfgset$}
\State Initialize $\prec$ to $\prec_\text{init}$, random non-repeating, numbers in $[0,1]$. \label{line:initial_priorities}
\While{some $\robot{i}$ not at any goal}
  \State $\mathcal{E}\gets\emptyset$ \Comment{Edge reservations (per step).}
  \State \textcolor{blue}{\textbf{for each} $\robot{i}$ in $\prec$ \textbf{do} \textsc{TrySwap}$(\robot{i}, g)$} \label{line:try-swap-loop} 
  \State \textbf{for each} $\robot{i}$ in $\prec$ \textbf{do} \textsc{TryMove}$(\robot{i})$ \label{line:try-move-loop}
  \State Commit $\mathcal{E}$; update $\robcfg{i}$; demote $\robot{i}$ at goals to $\prec_\text{init}$, increment others\label{line:commit}
\EndWhile
\State \Return $\robtrajset{}$

% \vspace{0.25em}
\Function{TryMove}{$\robot{i}$}
  \State \textcolor{black}{$\mathcal{E}^i \gets \mprimgen{i}(\robcfg{i})$; order by heuristic to $g(\robot{i})$} \label{line:gen-mprims}
  \For{each $e^i \in\mathcal{E}^i$} \Comment{Choose candidate primitives (i.e., edges).}
    \State \textcolor{black}{\textbf{if} $\exists e^j \in\mathcal{E}$ with edge--edge conflict, \textbf{continue}} \Comment{Primitive blocked} \label{line:edge-conflict}
    \State $U\!\gets\!\{\robot{j}\!\neq\!\robot{i}\mid \robot{j}\ \text{unreserved in }\mathcal{E}\}$ \Comment{Lower-priority.}
    \State \textcolor{blue}{$\mathcal{S}\!\gets\!\{\robot{j}\!\in\!U\mid e^i\ \text{edge--state conflicts }\robcfg{j}\}$} \label{line:lower-priority-stepped-on-set} \Comment{``Stepped-on'' set.}
    \If{\label{line:no-conflicts}$\mathcal{S}=\emptyset$} $\mathcal{E}\!\gets\!\mathcal{E}\cup\{e^i\}$; \Return \textbf{true} \EndIf
    \State \textcolor{blue}{$\mathcal{E}_{\rm curr}\gets\mathcal{E}$} \Comment{Record current edge reservations.}
    \State \textcolor{blue}{$\mathcal{E}\!\gets\!\mathcal{E}\cup\{e^i\}$} \Comment{Tentatively commit to edge.} % tentative reservation
    \For{\label{line:recursive-moves}\textcolor{blue}{each permutation $\pi$ of $\mathcal{S}$}} 
      \State \textcolor{blue}{$\mathrm{ok}\gets\textbf{true}$}
      \For{\textcolor{blue}{each $\robot{j}\in\pi$}}
        \If{\textbf{not} \textsc{TryMove}($\robot{j}$)} $\mathrm{ok}\!\gets\!\textbf{false}$; \textbf{break} \EndIf \label{line:try_move_permutation}
      \EndFor
      \State \textcolor{blue}{\textbf{if} {$\mathrm{ok}$} \textbf{then} \Return \textbf{true}} \Comment{All recursive calls succeeded.}
      \State \textcolor{blue}{\textbf{else} $\mathcal{E}\gets\mathcal{E}_{\rm curr}$} \label{line:rollback} \Comment{Roll back.}
    \EndFor
  \EndFor
  \State $e^i_{\text{wait}} \gets (\robcfg{i}\!\rightarrow\!\robcfg{i})$;\quad $\mathcal{E}\gets\mathcal{E}\cup\{e^i_{\text{wait}}\}$;\quad \Return \textbf{false} \label{line:wait-edge}
\EndFunction

\vspace{0.25em}
\Function{TrySwap}{$\robot{i}, g$}
  \For{\label{line:swap-loop}\textcolor{blue}{\textbf{each} $\robot{j}$ with $\robot{i}\prec \robot{j}$}} \Comment{Consider lower priority robots.}
    \State \textcolor{blue}{$d_{i,a}, d_{i,b} \gets \mathrm{dist}(\robcfg{i}, g(\robot{i})), \mathrm{dist}(\robcfg{i}, g(\robot{j}))$}
    \State \textcolor{blue}{$d_{j,a}, d_{j,b} \gets \mathrm{dist}(\robcfg{j}, g(\robot{j})), \mathrm{dist}(\robcfg{j}, g(\robot{i}))$}
    \If{\label{line:swap-condition}\textcolor{blue}{$d_{i,b} < d_{i,a}$  \Comment{Swap benefits higher-priority robot.} \\   \hspace*{0.5em} \textbf{ and } $ d_{i,b}+d_{j,b} \le d_{i,a}+d_{j,a}$}} \Comment{Benefits system.} \label{line:swap_if_conditional}
      \State \textcolor{blue}{swap $g(\robot{i}) \leftrightarrow g(\robot{j})$} \label{line:swap-goals}
      \State \textcolor{blue}{swap $\prec(\robot{i}) \leftrightarrow \prec(\robot{j})$} \label{swap-priorities}
    \EndIf
  \EndFor
\EndFunction
\end{algorithmic}
\end{algorithm}
\vspace{-0.32cm}

%% file: figures/fig_gco_manip_exps.tex
\begin{figure*}[t]
\centering

% ================= LEFT: main figure (will get its own Fig. #) =================
\begin{minipage}[t]{0.70\linewidth}
  \centering

  % Row 1
  \raisebox{0.7\height}{\includegraphics[height=0.7cm]{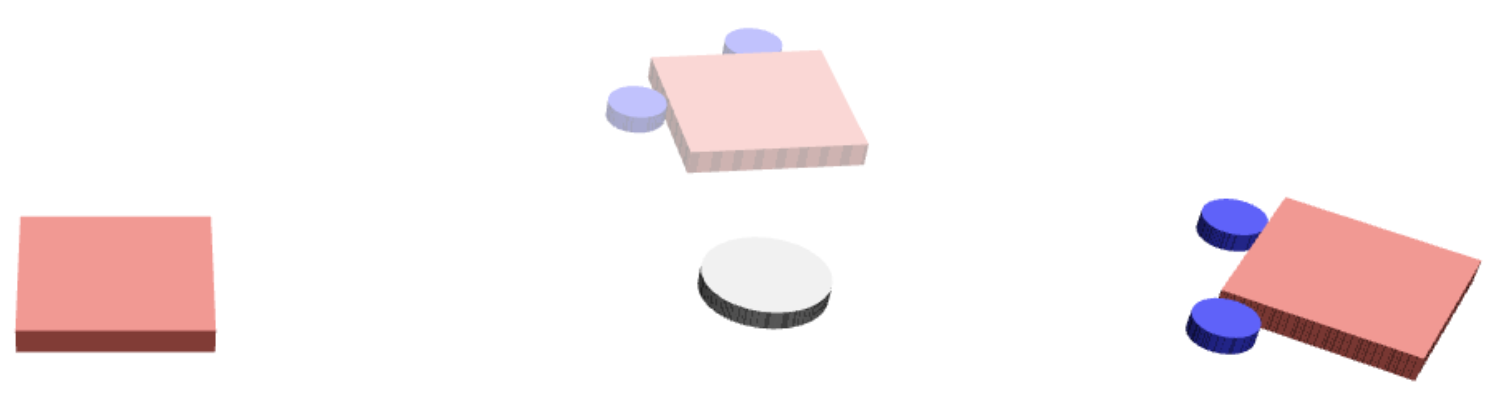}}%
  \hfill
  \includegraphics[height=1.4cm]{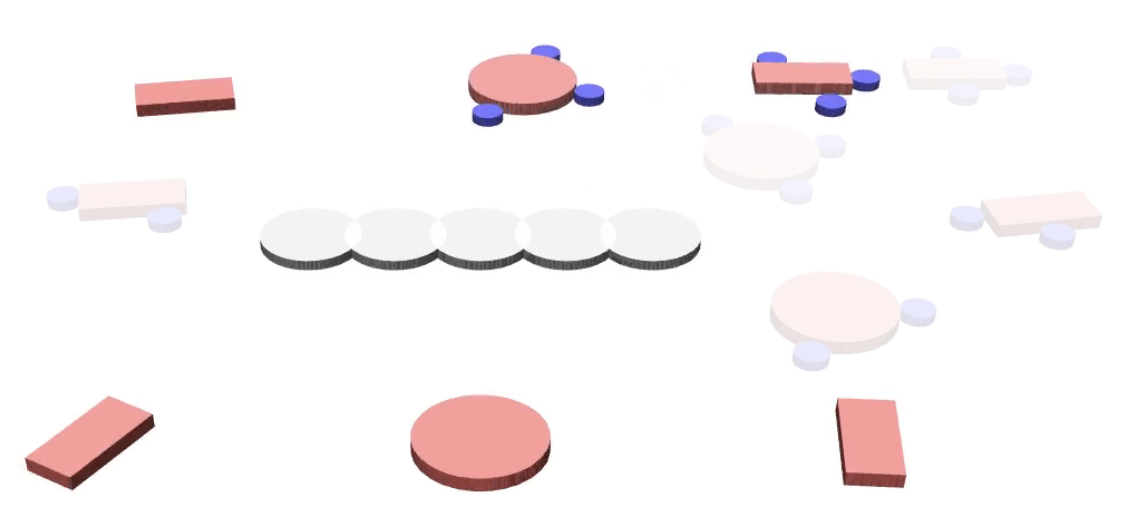}
  \includegraphics[height=1.6cm]{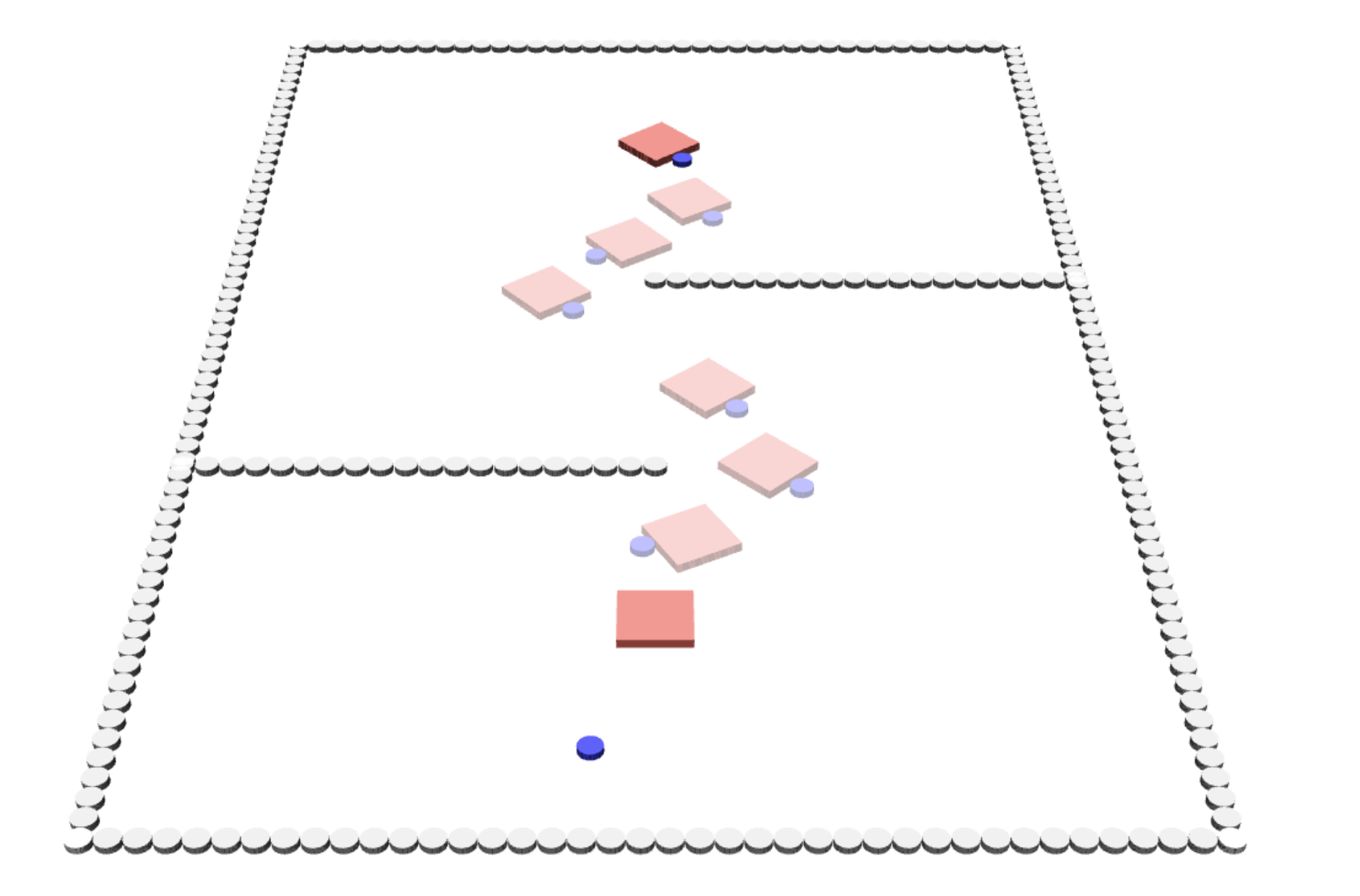}
  \includegraphics[height=1.4cm]{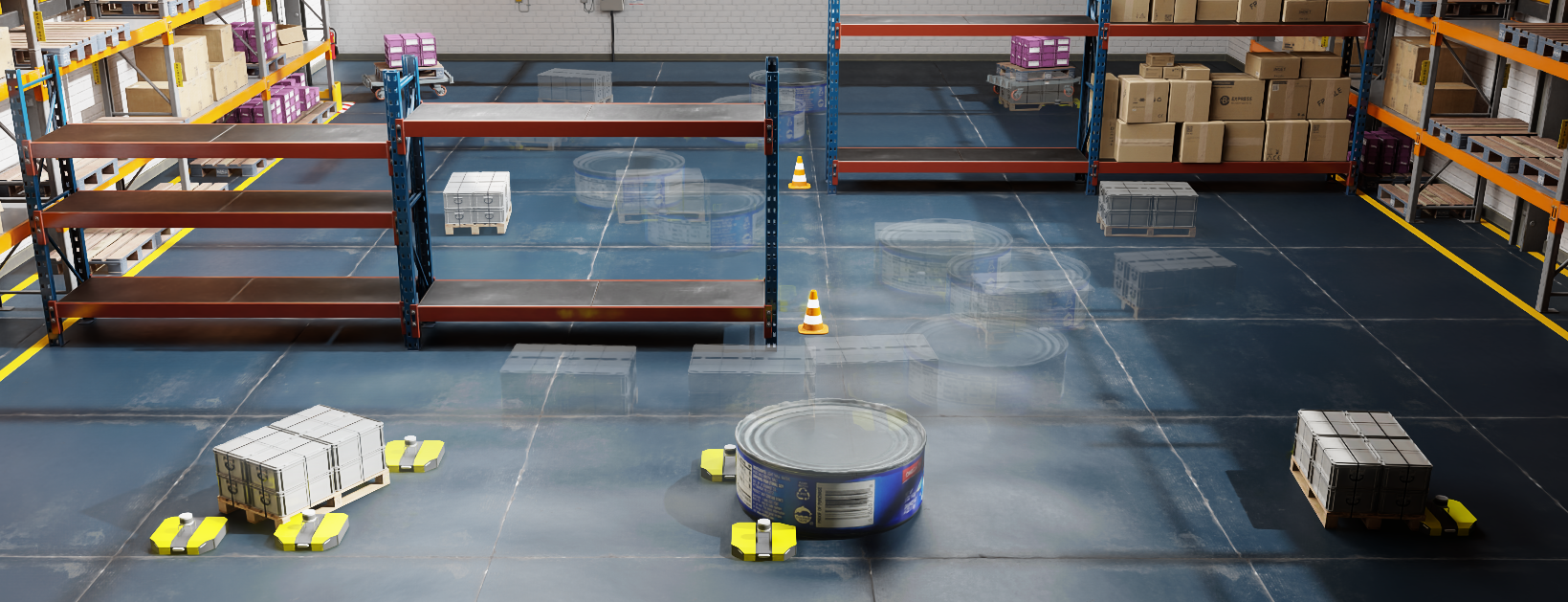}

  \vspace{0.3em}

  % Row 2
  \begin{minipage}[t]{0.29\linewidth}
    \includegraphics[width=\linewidth]{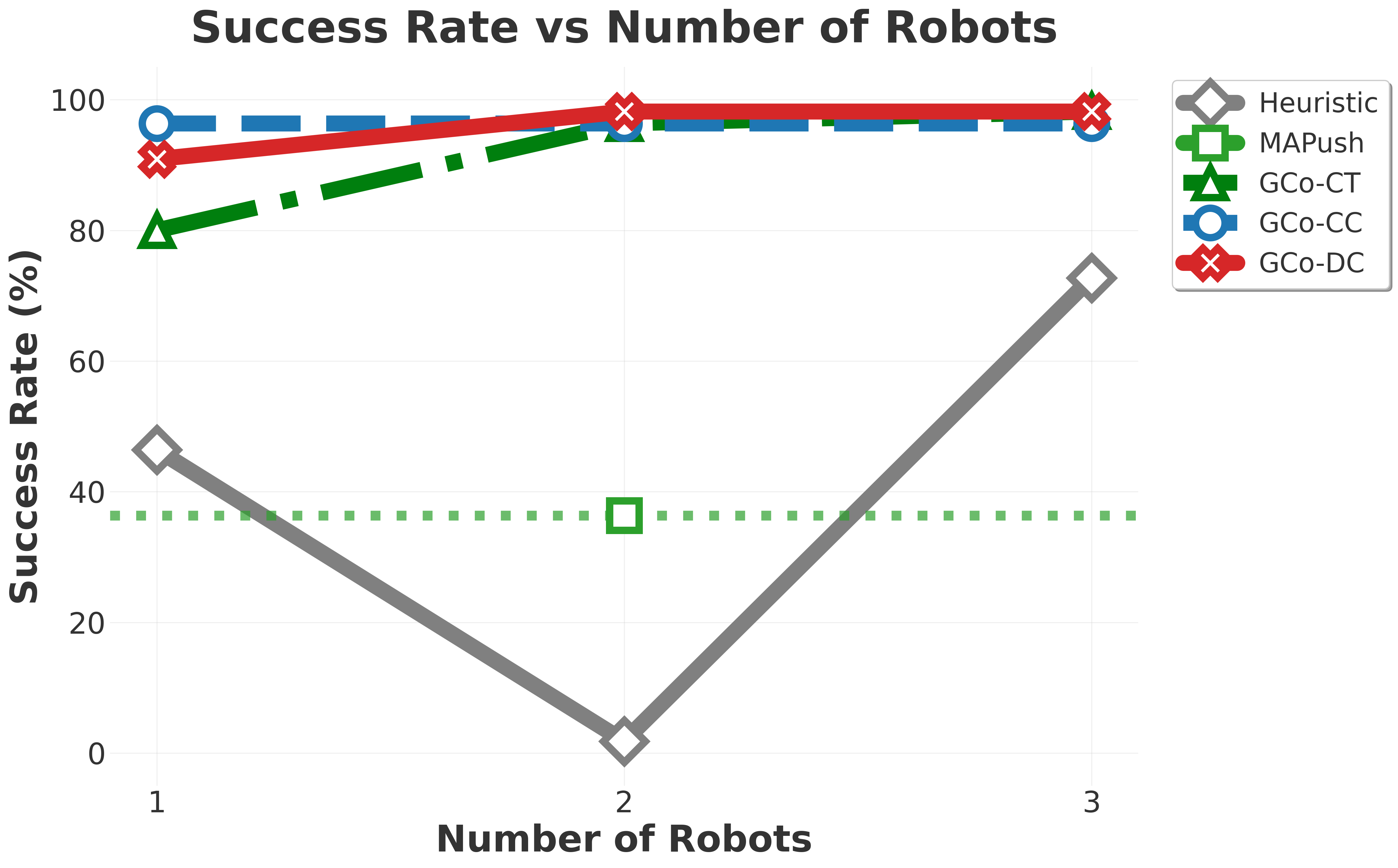}
  \end{minipage}\hfill
  \begin{minipage}[t]{0.29\linewidth}
    \includegraphics[width=\linewidth]{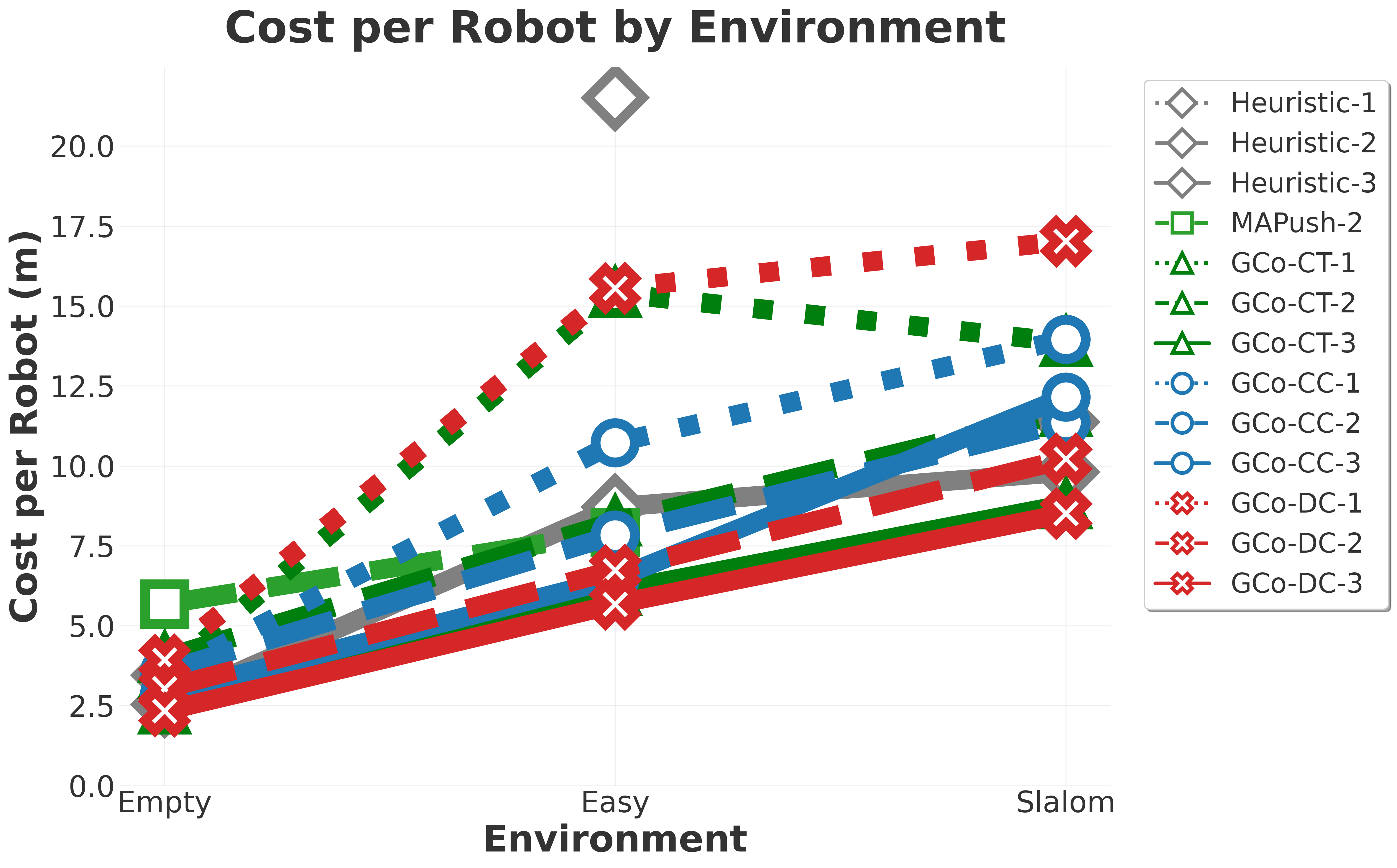}
  \end{minipage}\hfill
  \begin{minipage}[t]{0.4\linewidth}
    \includegraphics[width=\linewidth]{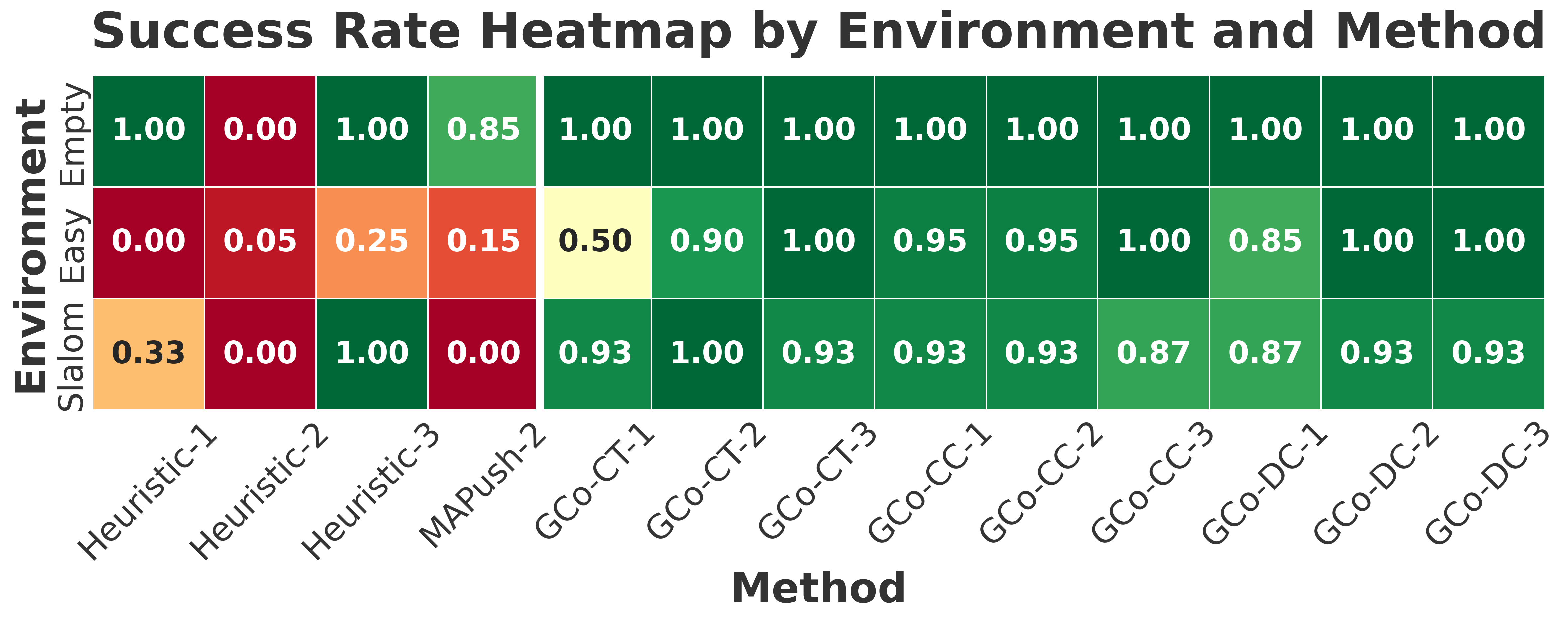}
  \end{minipage}

  % \vspace{0.3em}

  % Row 3
  \begin{minipage}[t]{0.29\linewidth}
    \includegraphics[width=\linewidth]{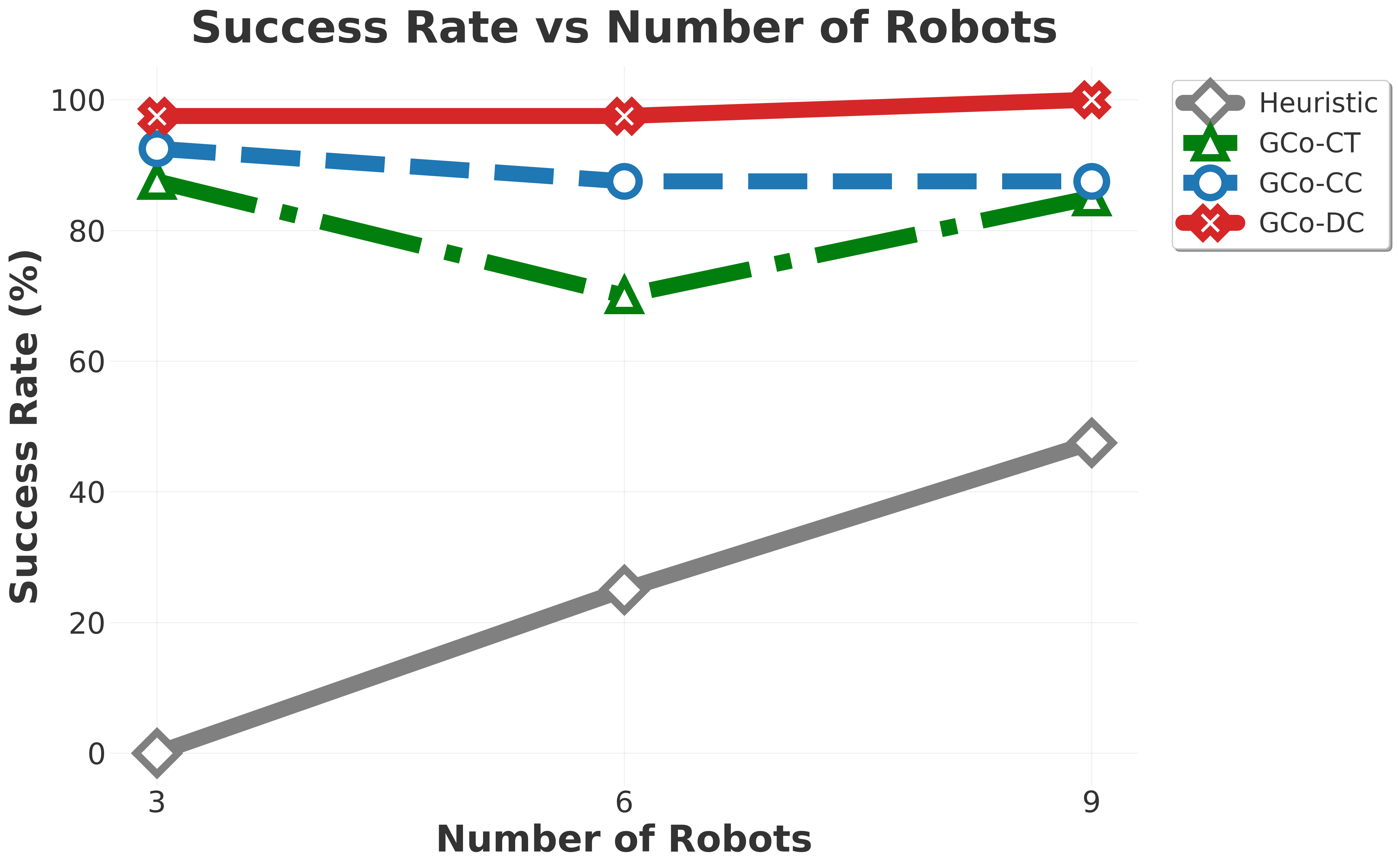}
  \end{minipage}\hfill
  \begin{minipage}[t]{0.29\linewidth}
    \includegraphics[width=\linewidth]{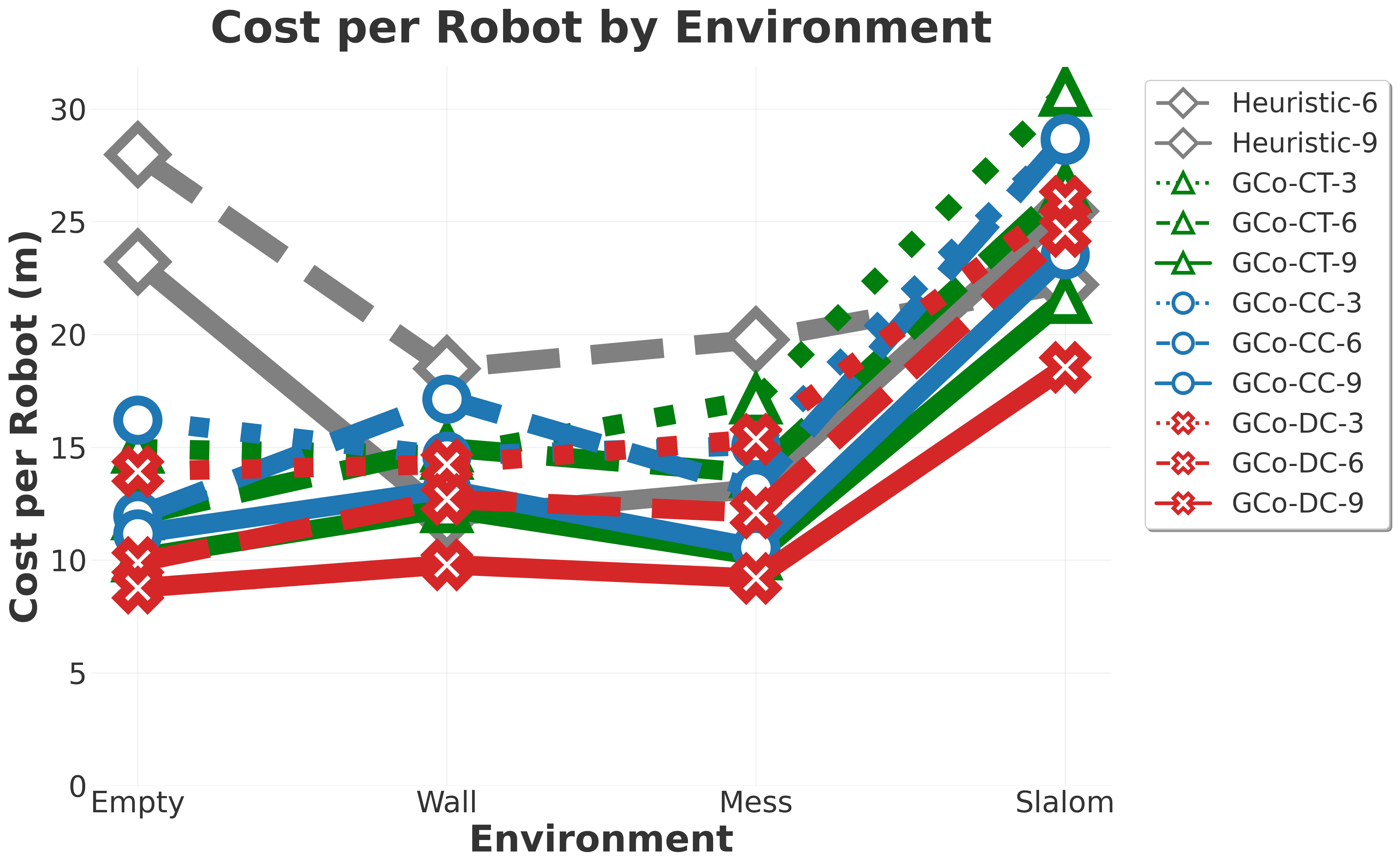}
  \end{minipage}\hfill
  \begin{minipage}[t]{0.4\linewidth}
    \includegraphics[width=\linewidth]{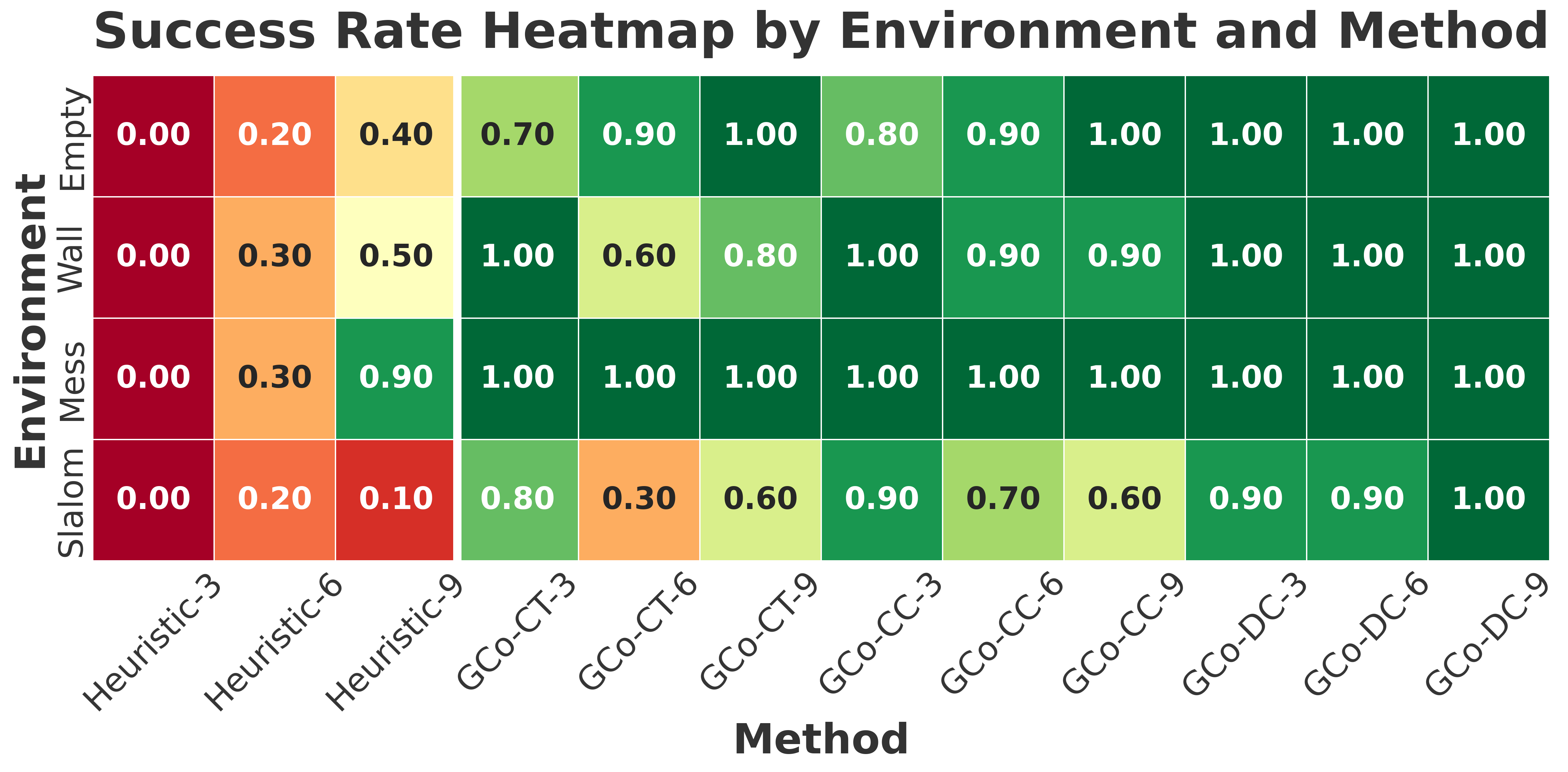}
  \end{minipage}

  \captionsetup{type=figure}
  \captionof{figure}{Multi-robot multi-object manipulation results.
  \gco{} consistently outperformed \textsc{MAPush} and \textsc{Heuristic} baselines with \gcodc{} performing the best. Top row: \textit{Easy}, \textit{Wall}, and \textit{Slalom} (single and multi-object versions) setups.
  Middle and bottom rows: single- and multi-object manipulation, respectively. We report overall success rates (left column), overall average distance traveled per robot (middle), and a breakdown of success rates. Number in method names is the number of robots available in the scene. } \label{fig:all_results}
\end{minipage}
\hfill
% ================= RIGHT: thin figure (will get its own Fig. #) =================
\begin{minipage}[t]{0.28\linewidth}
  \centering
  
  \includegraphics[height=1.6cm]{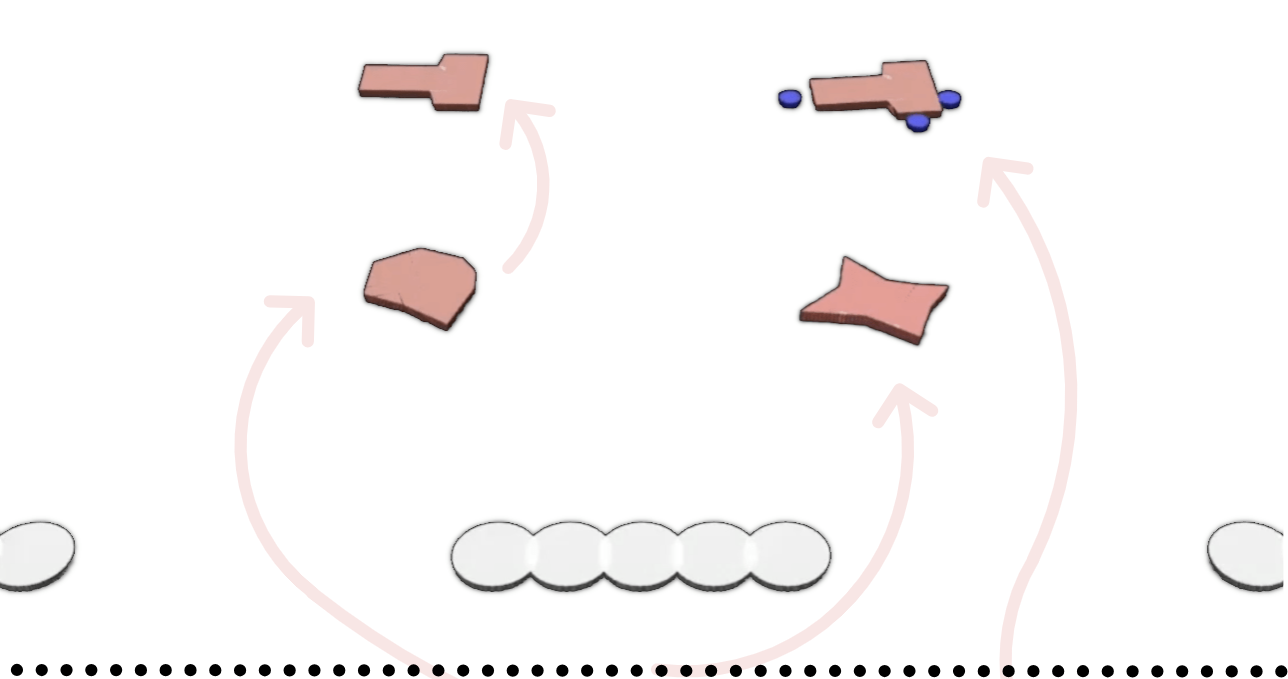}

  % \vspace{0.35em}

  \includegraphics[width=0.8\linewidth]{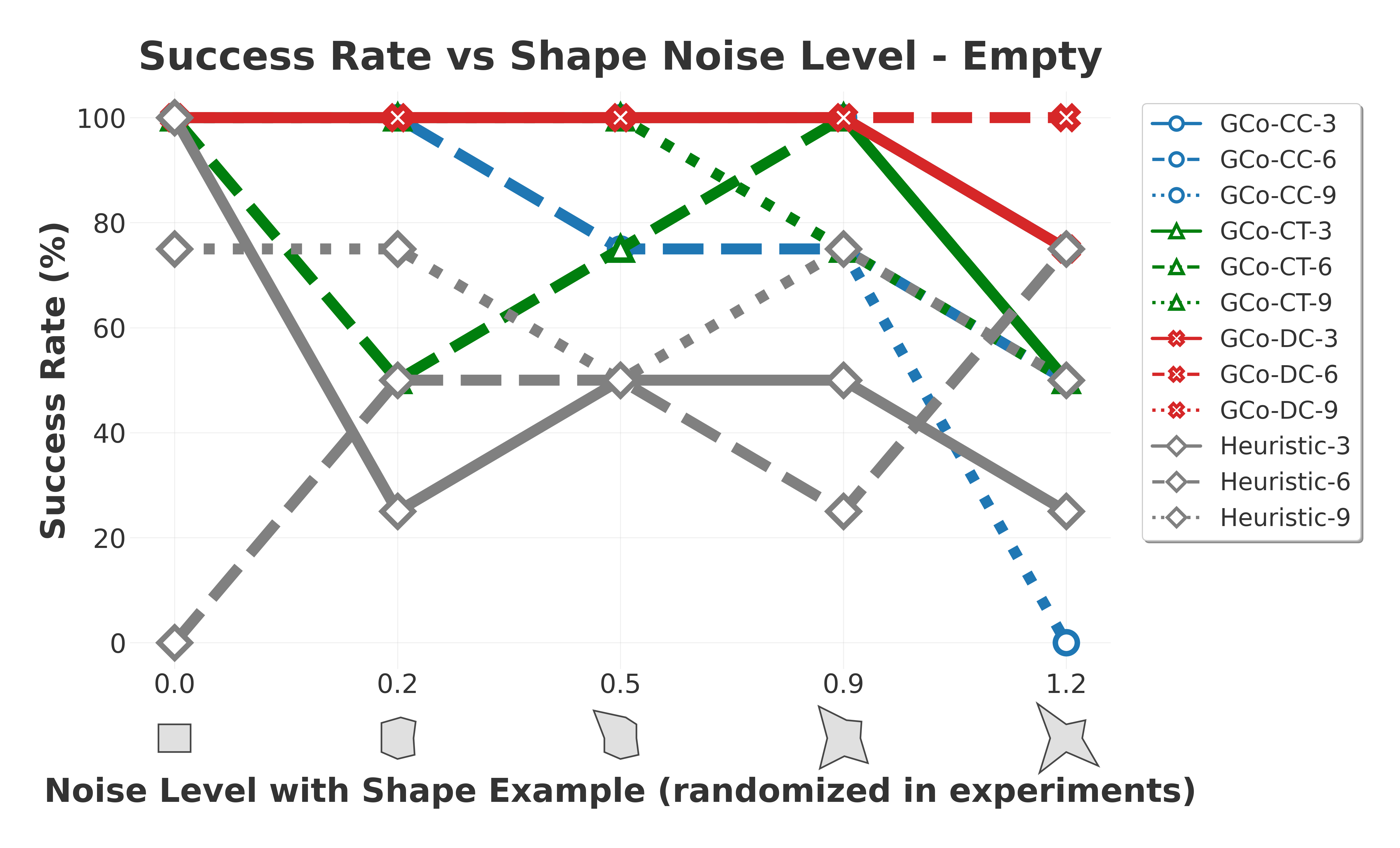}

  % \vspace{0.35em}

  % replace with the multi-object OOD plot if different
  \includegraphics[width=0.8\linewidth]{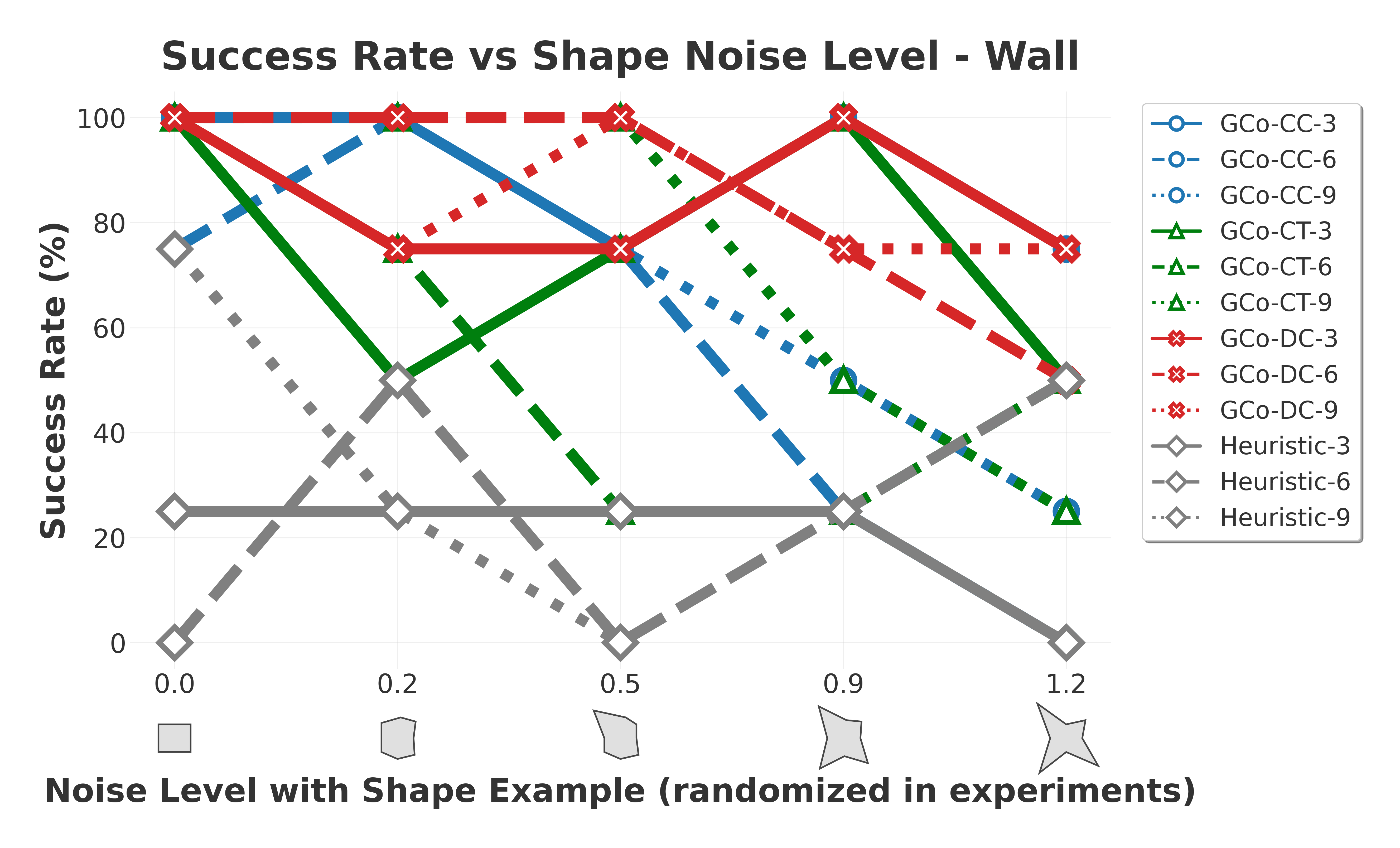}

  \captionsetup{type=figure}
  \captionof{figure}{Robustness to shape perturbations. Top: three robots manipulating objects drastically different from their training set. Middle and bottom: success rates in the \textit{Empty} and \textit{Wall} maps, both with 4 objects.} \label{fig:ood_results}
\end{minipage}
\vspace{-0.7cm}
\end{figure*}

%% file: tex/experiments.tex
\input{figures/fig_amrmp_exps}
\section{Experimental Evaluation}
\label{sec:experiments}

We evaluated \gco{} and \gspi{} across a diverse set of manipulation and motion-planning scenarios designed to probe performance in increasing coordination difficulty, horizon length, scale, and object complexity.

% ====== Manipulation Experiments ======
\subsection{Manipulation Experiments}
\label{sec:exp_manip}
We constructed 300 problems spanning three regimes: single-object manipulation, enabling direct comparison with HRL baselines (Fig.~\ref{fig:all_results}, middle row); multi-object manipulation, testing scalability and inter-object coordination (Fig.~\ref{fig:all_results}, bottom row); and out-of-distribution (OOD) scenarios, evaluating robustness to novel objects drastically different from those seen in training (Fig. \ref{fig:ood_results}). Success required all objects within 15\,cm and 0.5\,rad of targets, and cost was the average distance traveled per robot. Single-object problems were repeated $10$ times, with the rest repeating $5$ times.

\subsubsection{Evaluated Methods}
We included all \gco{} variants, trained on a shared dataset.\footnote{Our training set contained $20{,}000$ samples of variable-sized boxes and cylinders. Data were collected in MuJoCo~\cite{todorov2012mujoco} by simulating teams of varying sizes executing diverse contact formations and recording outcomes.}
We compared \gco{} against \textsc{MAPush}~\cite{feng2025mapush}, a recent reinforcement-learning method for multi-agent single-object pushing\footnote{We used the official implementation and provided 2-robot models. Additionally, we simplified the problems given to \textsc{MAPush} to resemble those it has been trained on: we allowed a translation tolerance of $0.5$\,m and ignored orientation errors (github.com/collaborative-mapush/MAPush).},
and a geometric baseline inspired by~\cite{chen2015occlusion} referred to as \textsc{Heuristic}, which mirrors the \gco{} pipeline but replaces the learned policy $\pi_\theta$ with a hand-designed contact and trajectory generator. Succinctly, given an object $\obj{j}$ and requested transformation $T^j$, \textsc{Heuristic} draws evenly spaced lines (as many as the budget allows) from the object's centroid against its requested direction of travel and selects contact points where these lines intersect with the object's boundary; emulating execution-time robot motion to contact. \textsc{Heuristic} generates manipulation trajectories by applying $T^j$ to the contact points and linearly interpolating between the original and transformed points. Robot radii were 0.1 meters.
% If a contact overlaps the object, the point (and associated trajectory) is shifted radially outward from the object centroid.

\subsubsection{Single-Object Manipulation Problems}
In the simplest regime, we considered a single square object and robot teams of one to three.
We conducted our experiments in three maps: \emph{Empty}, \emph{Easy}, and \emph{Slalom}.
In \emph{Empty} and \emph{Easy}, robots were asked to translate the object by 1\,m without rotation or by 2\,m while rotating by 2\,rad, respectively.
In both settings, we requested motions along $\pm x$ and $\pm y$; \emph{Empty} was obstacle-free, while \emph{Easy} contained a single obstacle between start and goal.
In \emph{Slalom}, the object was asked to reach the other side of the maze-like environment from bottom to top, emphasizing long-horizon manipulation (Fig.~\ref{fig:all_results}).

\subsubsection{Multi-Object Manipulation Problems}
Scaling up, we requested teams of $3$, $6$, or $9$ robots to manipulate up to five objects with randomized shapes and sizes over translations of up to six meters and rotations of roughly $2.5$ radians.
For our tests, we reused \textit{Empty} and \textit{Slalom} maps, this time with more objects, and added two new settings: \textit{Wall} and \emph{Mess}. In the \textit{Empty} and \textit{Wall} maps, up to five objects were placed in a row and were required to move to a parallel row of goals 2\,m away. \emph{Wall} had a wide obstacle blocking the center. In the \textit{Mess} setting, four scattered objects were required to be arranged in a square formation, and in the \emph{Slalom} map, objects were asked to traverse tight corridors in a maze, where manipulation errors could result in blocked progress. In all problems, we shuffled the initial object–goal assignment, requiring goal swaps for efficient solutions.

\subsubsection{Out-Of-Distribution Problems} Despite being trained only on boxes and cylinders, we evaluated \gco{} on general polygon and ``T-shaped'' objects. Polygons with noise $\sigma$ were generated by radially shifting vertices on a rectangle (placed mid-edges and on corners) by $\text{ReLU}(\epsilon)$, where $\epsilon \sim \mathcal{N}(0, \sigma)$.

\subsubsection{Manipulation Results}
Overall, our results indicate that \gco{} scales well to multi-robot, multi-object manipulation and that \gcodc{} is consistently the most stable as task difficulty increases.
In single-object tasks (Fig.~\ref{fig:all_results}, middle row), all \gco{} variants achieved higher success rates and lower costs than the baselines across most settings. \textsc{MAPush} performed well in the simplest case but quickly degraded on longer-horizon problems. \textsc{Heuristic} was inconsistent and sensitive to team size, with failures often caused by brittle contact choices that led to slip. In contrast, \gco{} solved nearly all problems, with cost per robot decreasing with team size, suggesting efficient use of additional robots.
In multi-object tasks (Fig.~\ref{fig:all_results}, bottom row), a clear hierarchy emerged: \gcodc{} achieved the highest success rates and lowest costs, with \gcocc{} and \gcoct{} trailing but still outperforming the heuristic baseline; \textsc{MAPush} was not applicable in this regime. 
% Since the \gco{} variants differ only in contact and trajectory parameterization, this gap highlights the stability benefits of discrete contact selection combined with continuous trajectory generation.
In OOD tests, despite being trained only on boxes and cylinders, \gco{} remained robust on deformed shapes, with \gcodc{} showing a slow degradation as noise increased (Fig.~\ref{fig:ood_results}). 
We allowed methods long computation limits to evaluate only their manipulation capacity: 100 iterations for \gco{} and \textsc{Heuristic}, and 2 simulation minutes for \textsc{MAPush}.
Overall, we find these manipulation results especially promising since our problem sets extend the difficulty of common prior settings in object count and shape, team size, horizon length, and obstacles \cite{feng2025mapush, xiong2024mqe, nachum2019multi, jeon2023learning}.
\subsection{\gspi{} AMRMP Experimental Analysis}
\label{sec:exp_amrmp}

To evaluate \gspi{} in isolation from \gco{}, we constructed 274 AMRMP tasks with up to 125 robots spanning three regimes: obstacle-free environments for large-scale tests (Fig.~\ref{fig:gspi_all_results}, top plots and illustrations), obstacle-dense maps requiring tight coordination (Fig.~\ref{fig:gspi_all_results}, bottom plots and middle illustrations), and targeted stress tests designed to expose common failure modes (Fig.~\ref{fig:gspi_all_results}, bottom left illustrations). A run was successful if all robots safely reached goals; solution cost was the sum of travel distances. Problems were repeated five times with random initial robot-goal assignments.

\subsubsection{Evaluated Methods}
We compared \gspi{} to state-of-the-art and adapted baselines. The primary comparator was C-UNAV~\cite{dergachev2024dec-unav}, which uses ORCA~\cite{van2011orca} for navigation and a goal-swapping mechanism.\footnote{We implemented C-UNAV by adding goal-swapping to the ORCA code from the same authors (github.com/PathPlanning/ORCA-algorithm).}
We also included TSWAP and PIBT, adapted to operate over motion primitives rather than grid edges. \gspi{}, PIBT, and TSWAP used the same primitive set: short (5\,cm) motions in eight cardinal directions. \gspi{}, C-UNAV, and TSWAP leverage goal swaps, while PIBT does not; we included PIBT due to its algorithmic similarity to \gspi{} to isolate the contribution of goal swapping within a PIBT-like structure. As all methods commit one multi-robot move per iteration, we also report \emph{iteration time}, defined as the average compute time required to commit to a move.

\subsubsection{Freespace AMRMP Problems}
Seeking to quantify \gspi{}'s ability to scale to large team sizes, we created three obstacle-free problem families with up to 125 robots. Our setups included transitions between formations where a high degree of symmetry and proximity between robots arose.

\subsubsection{Cluttered AMRMP Problems}
For evaluating coordination among obstacles and bottlenecks, we created three additional maps with team sizes up to 30 robots. These include formation transitions around obstacles, a slalom-style map that forces robots through a narrow passage, and a funnel-like variant that further tightens the bottleneck. 
% Negotiating a heavily contested shared space was necessary in these problems.

\subsubsection{Stress-Test AMRMP Problems}
To probe common failure modes under tight coupling and constrained geometry, we created five targeted stress-test problems (illustrated in Fig.~\ref{fig:gspi_all_results}, bottom left). These included tight layouts with densely packed goals, where robots needed to coordinate group motions to progress. We also included an enclosed map where starts and goals overlap, but robots cannot move. They must exchange goals in order to find a solving assignment. 

\subsubsection{AMRMP Results}
\gspi{} consistently solved more instances than competing methods while achieving lower or comparable solution costs. PIBT frequently failed due to livelocks, and TSWAP often deadlocked, particularly in symmetric settings. C-UNAV performed better in open space but struggled in scenarios with higher robot- or obstacle-density. This behavior persisted despite substantial ORCA parameter tuning. In contrast, \gspi{} remained robust across all tested scales, including teams exceeding 100 robots. Qualitatively, the trajectories in Fig. \ref{fig:gspi_all_results} that show robots rarely crossing paths indicate that \gspi{} performed effective goal swapping and quickly improved on the initial random robot-goal assignments. Otherwise, trajectories would have been intertwined.
Finally, the stress tests were particularly discriminative. \gspi{} solved 100\% of the 80 runs, whereas TSWAP solved 12.5\%, PIBT solved 2.5\%, and C-UNAV solved 0\%. These results suggest that \gspi{} not only scales well and effectively negotiates heavily contested regions, but also circumvents common failure modes.  

% \subsection{Implementation Details} 
% We implemented \gco{} in Python with PyTorch, with only \gspi{} in \texttt{C++} and exposed with Python bindings. Our flow matching implementations build on the code from \cite{lipman2024flowmatchingguidecode}. 
% We used $B_{\max{}}=3$ in all of our models, with training data including examples with $1$, $2$, and $3$ robots per object.
% All training and experiments were done on a laptop with an Nvidia RTX 3080-Ti GPU and an Intel Core i9-12900H with 32GB RAM (5.2GHz). Our code will be made public.

% If accepted, our code will be made available online at \href{https://gco-paper.github.io}{\texttt{gco-paper.github.io}}.

%% file: figures/fig_amrmp_exps.tex
\begin{figure*}[t]
    \centering
    \vspace{1mm}
    \begin{minipage}{0.35\linewidth}
        \includegraphics[width=\linewidth]{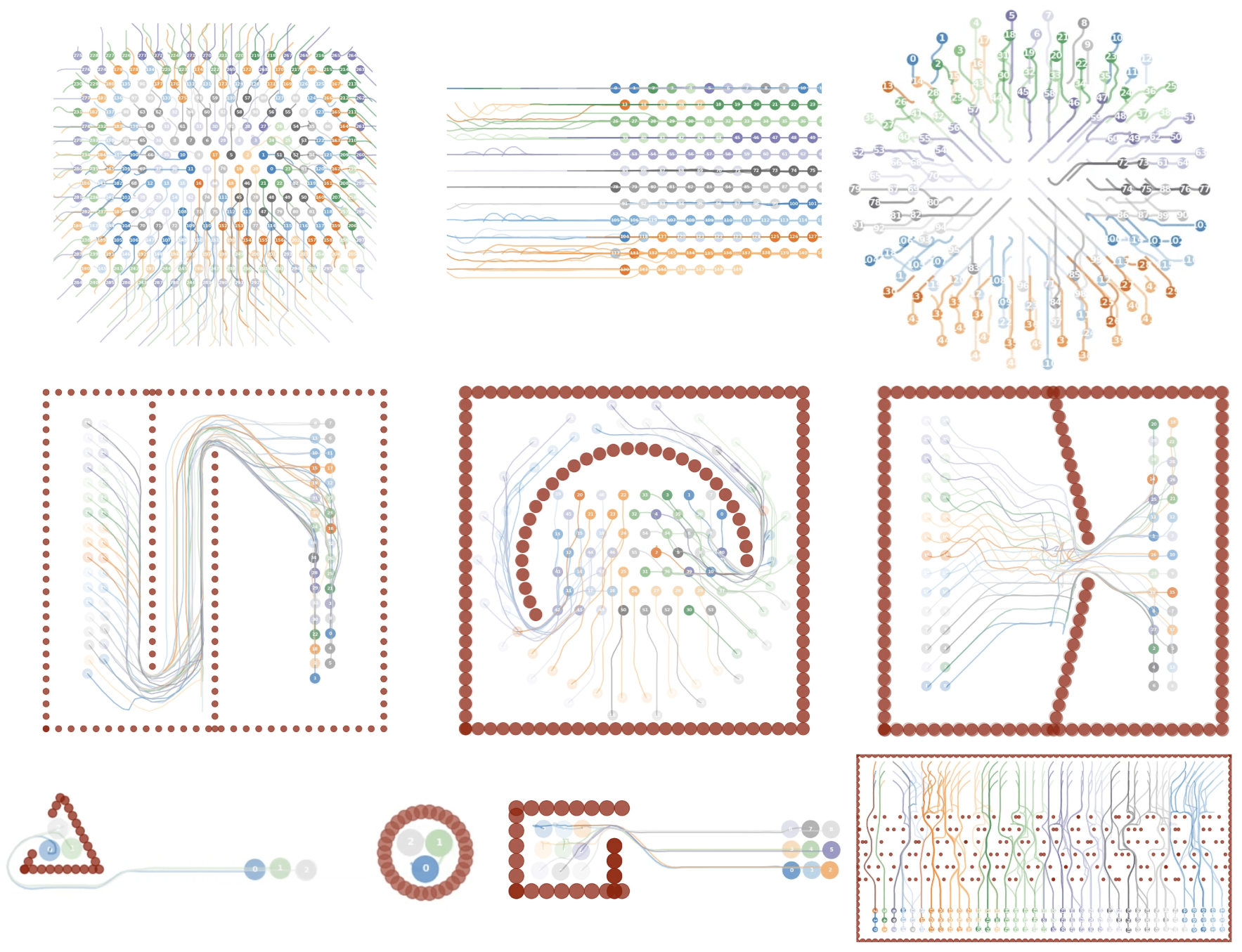}
    \end{minipage}
    \begin{minipage}{0.64\linewidth}
        \begin{minipage}{0.33\linewidth}
            \includegraphics[width=\linewidth]{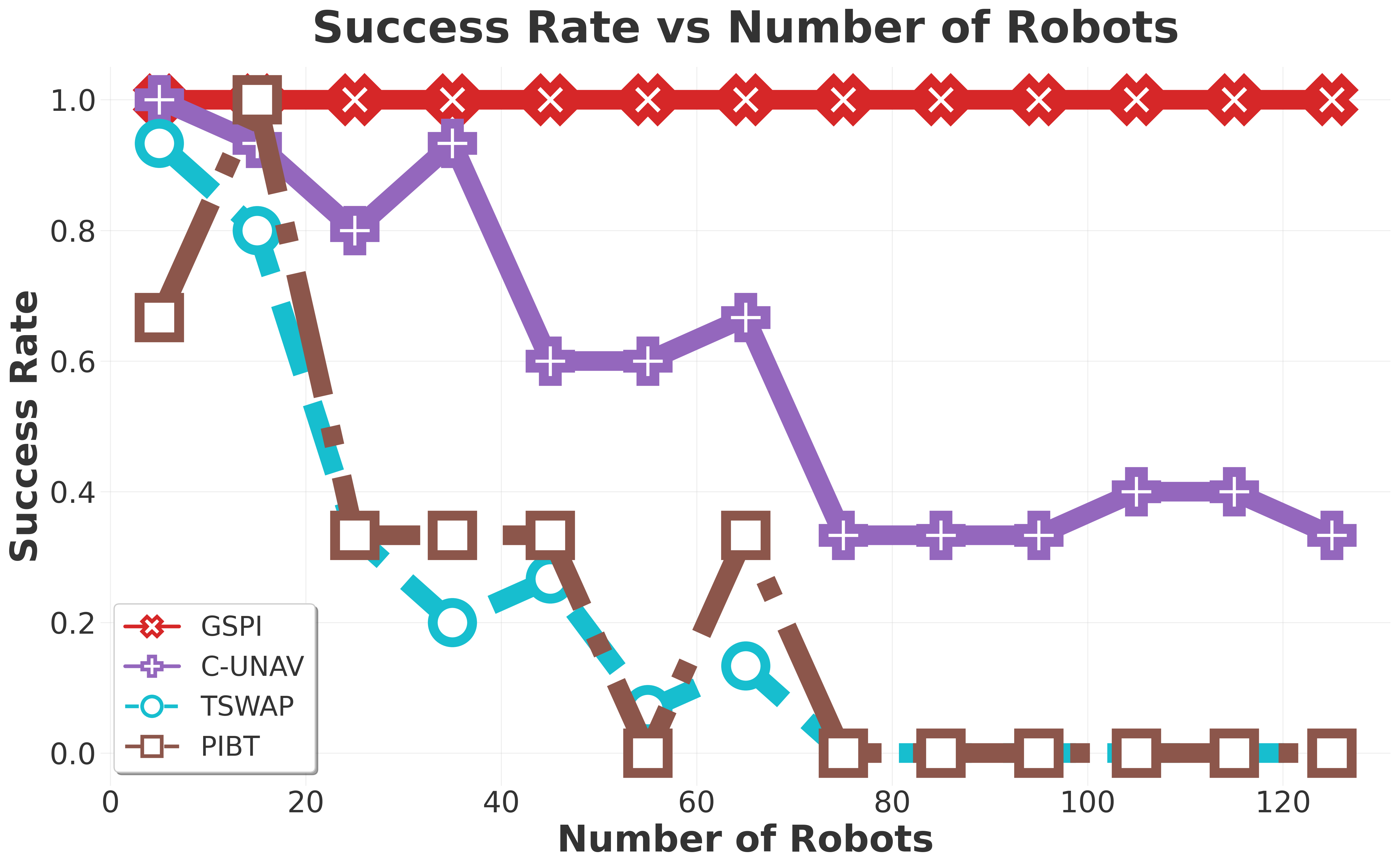}
        \end{minipage}
        \begin{minipage}{0.33\linewidth}
            \includegraphics[width=\linewidth]{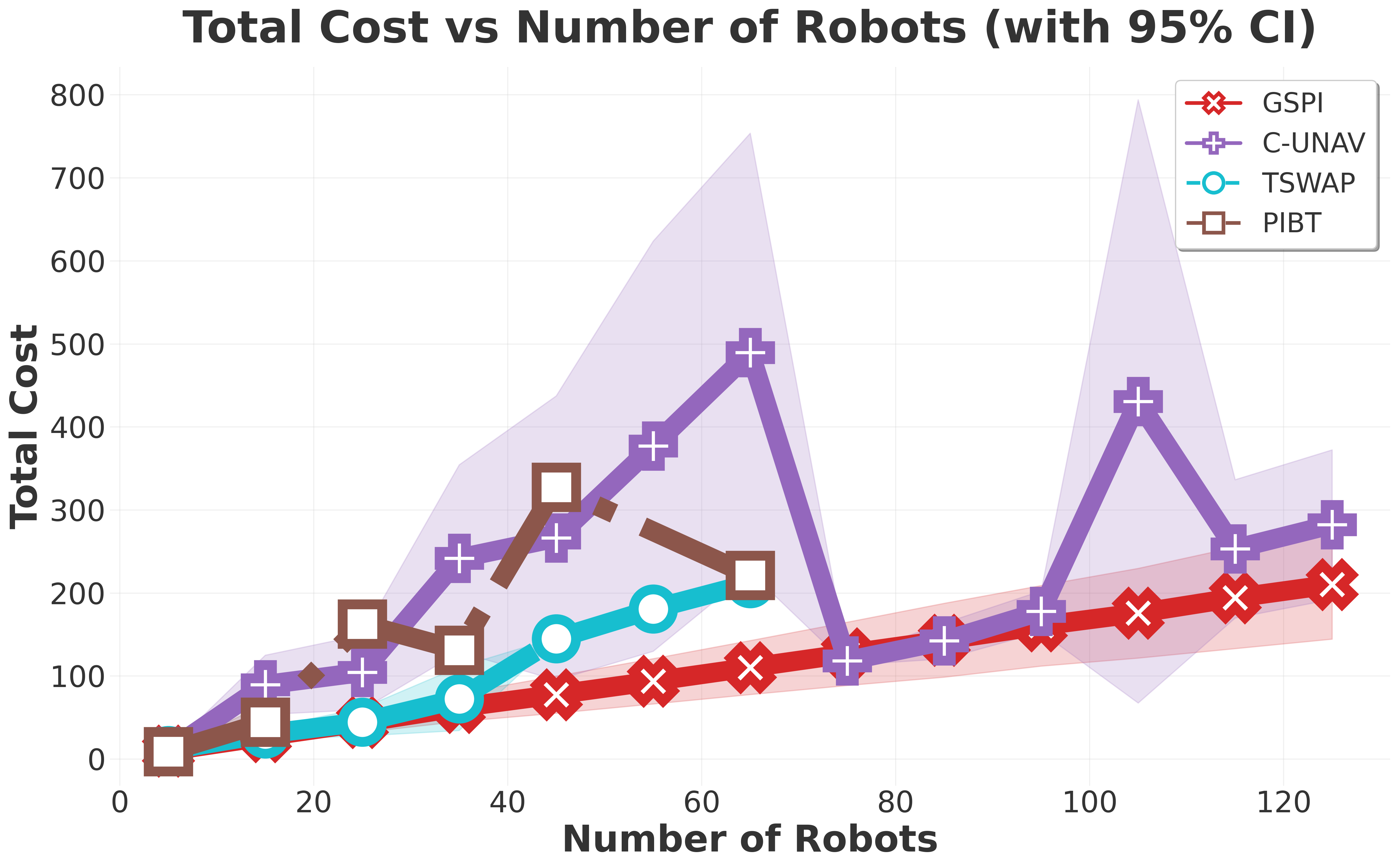}
        \end{minipage}
        \begin{minipage}{0.32\linewidth}
            \includegraphics[width=\linewidth]{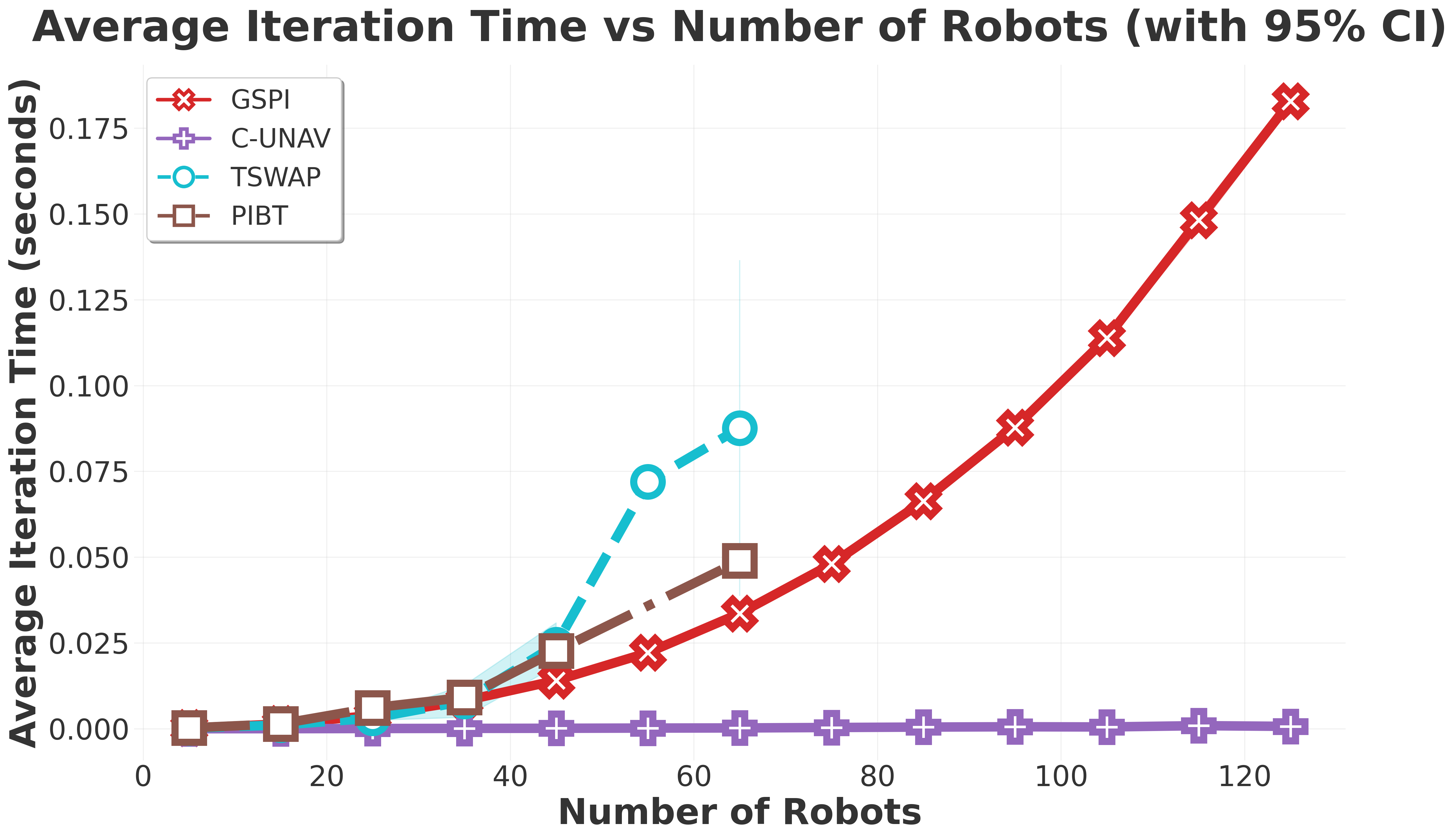}
        \end{minipage}
    % \vspace{1em} % Space between rows
    % Second row
        \begin{minipage}{0.33\linewidth}
            \includegraphics[width=\linewidth]{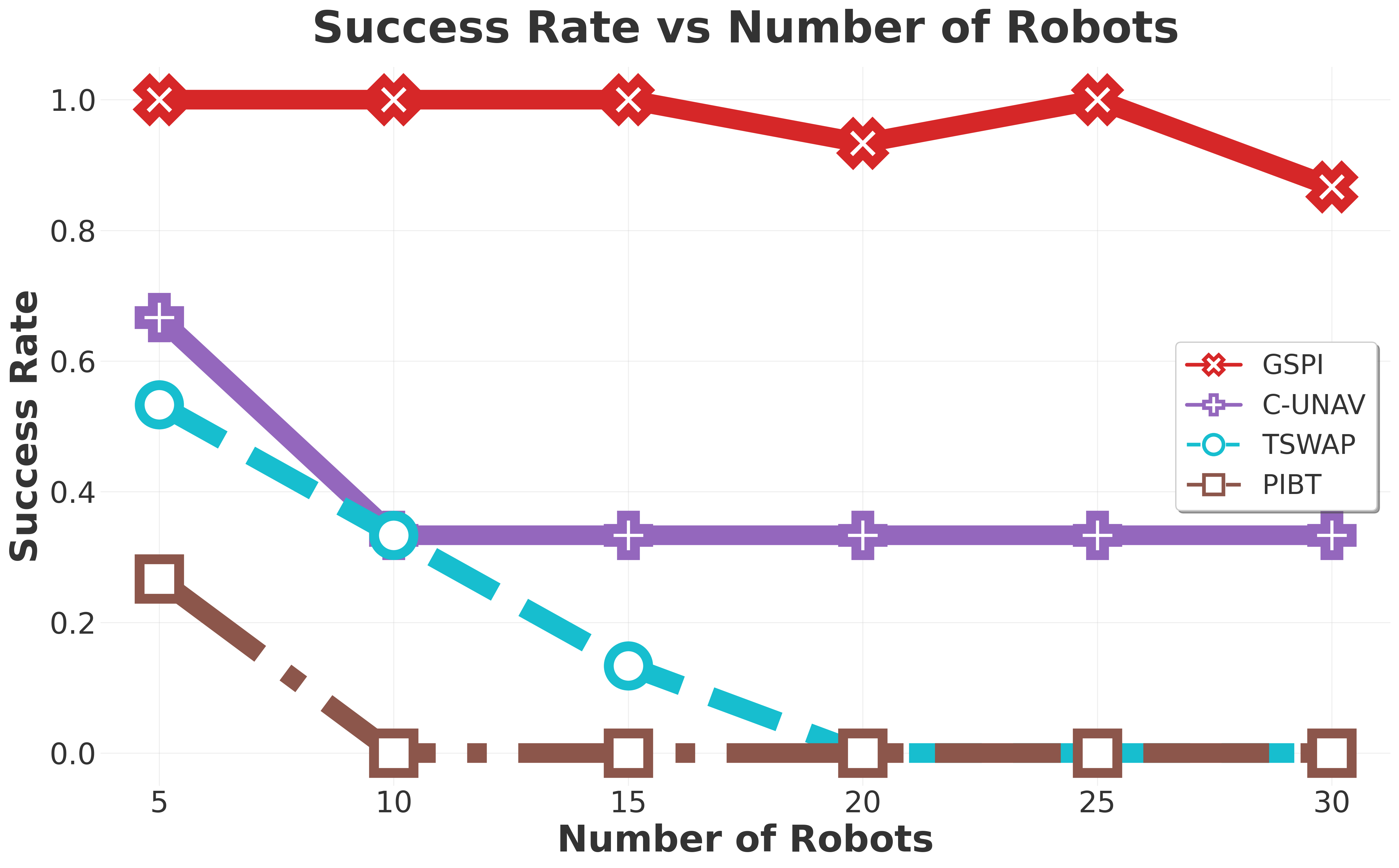}
        \end{minipage}
        \begin{minipage}{0.33\linewidth}
            \includegraphics[width=\linewidth]{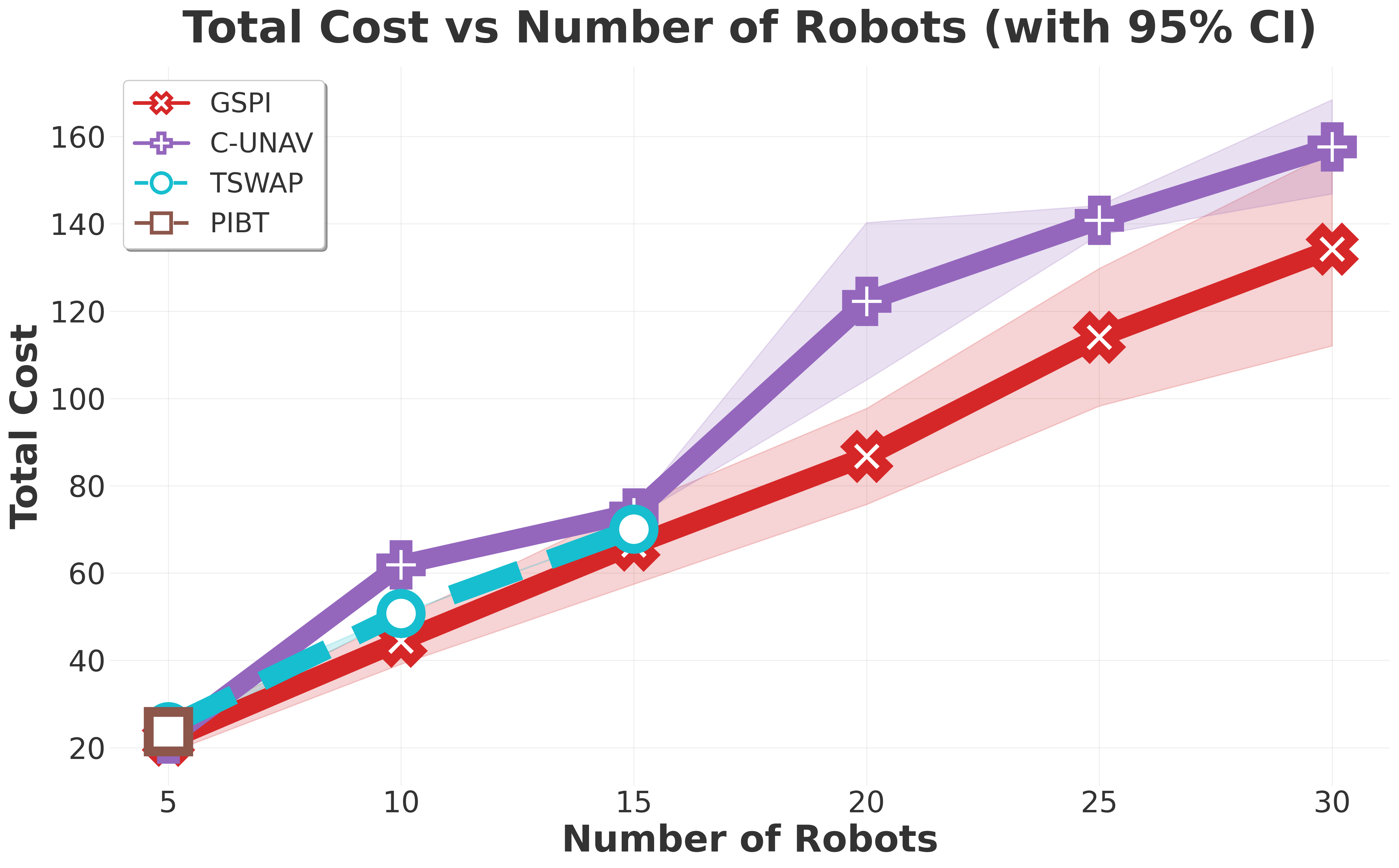}
        \end{minipage}
        \begin{minipage}{0.32\linewidth}
            \includegraphics[width=\linewidth]{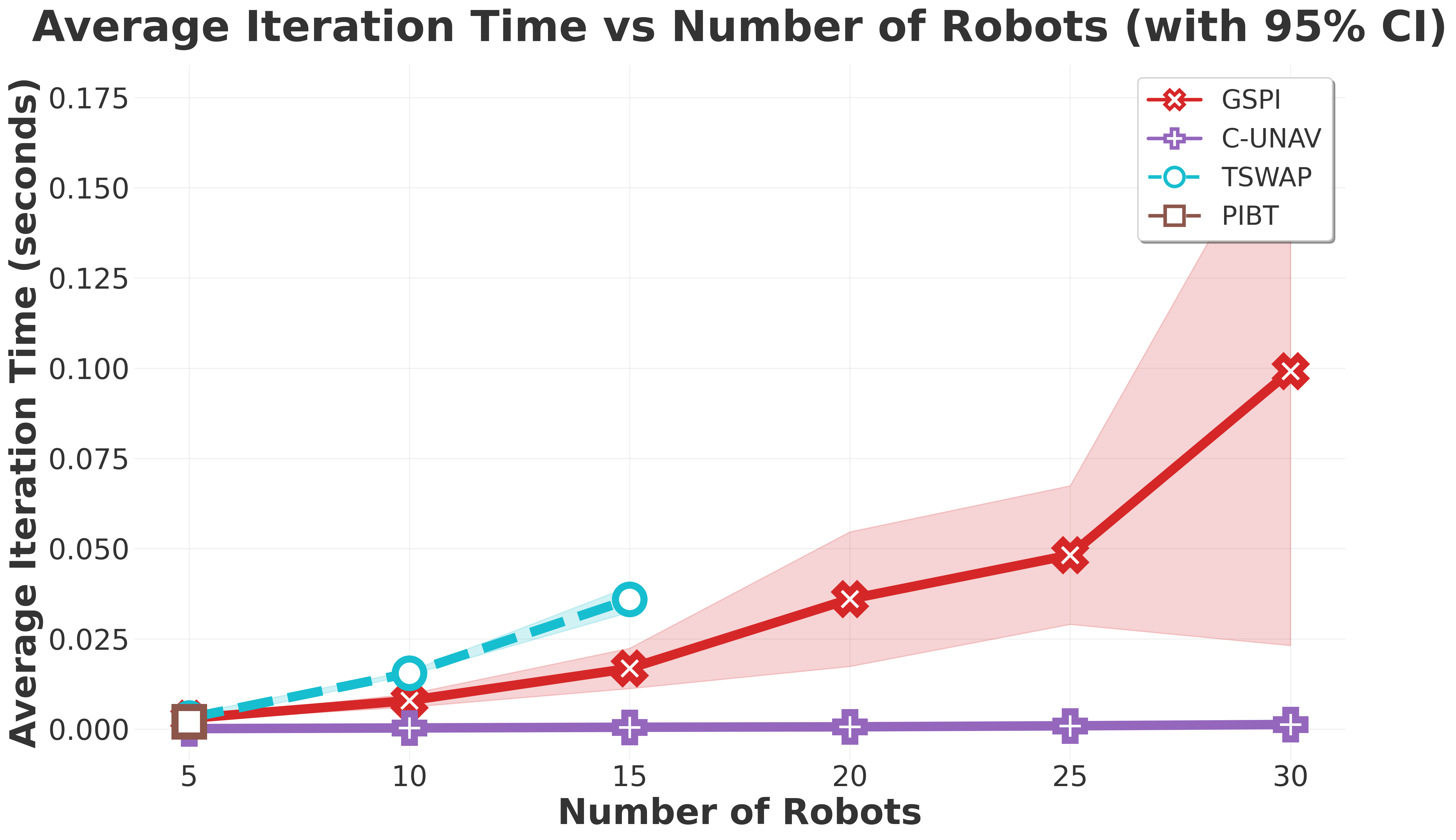}
        \end{minipage}
    \end{minipage}
    \vspace{-0.1cm}
    \captionsetup{belowskip=-10pt}
    \caption{
    \gspi{} analysis. Illustrations on the left show planned trajectories for all robots in freespace (top row), cluttered (middle row), and stress-test (bottom row) scenarios. The plots show success rates, costs, and iteration times in freespace (top row) and cluttered (bottom row) scenarios. \gspi{} solved significantly more problems than baselines with on-par or better cost. 
    % Its iteration time increasing with problem size but remained short.
    }
    \label{fig:gspi_all_results}
    \vspace{-0.3cm}
\end{figure*}

%% file: tex/conclusion.tex
\section{Conclusion}
\label{sec:conclusion}

We introduced \gco{}, a unified framework for collaborative multi-robot, multi-object non-prehensile manipulation that combines the strengths of generative modeling and multi-robot motion planning. At its core, \gco{} leverages flow matching co-generation to propose contact formations and manipulation trajectories directly from perception, and integrates them with \gspi{}, a new scalable algorithm for anonymous multi-robot motion planning. This enables robots not only to decide \emph{how} to interact with objects, but also to coordinate efficiently across large teams and dense clutter.

Through extensive experiments, we demonstrated that \gco{} consistently outperformed reinforcement learning and heuristic baselines in single- and multi-object manipulation, and that the discrete–continuous instantiation delivered the best performance. Moreover, we showed that \gspi{} achieved substantial scalability in AMRMP, successfully solving dense problems with more than one hundred robots where prior approaches struggle. In this light, we find that generative co-design, coupled with lightweight planning, is a powerful recipe for collaborative manipulation.

% Looking ahead, our framework opens several directions for future work. Incorporating richer perception modalities could extend applicability to diverse and deformable objects. Tightening the theoretical guarantees of \gspi{} would strengthen its role as a reliable backbone for large-scale systems. Finally, closing the loop with real-world deployments will test \gco{} under the uncertainties of sensing, actuation, and unmodeled dynamics. We believe that these steps will bring multi-robot collaboration closer to practical, scalable autonomy in complex environments.